\documentclass{article}

\usepackage{amsmath,amsfonts,bm}

\def\eqref#1{equation~\ref{#1}}

\def\1{\bm{1}}

\DeclareMathAlphabet{\mathsfit}{\encodingdefault}{\sfdefault}{m}{sl}
\SetMathAlphabet{\mathsfit}{bold}{\encodingdefault}{\sfdefault}{bx}{n}

\usepackage{PRIMEarxiv}

\usepackage[utf8]{inputenc} 
\usepackage[T1]{fontenc}    
\usepackage{hyperref}       
\usepackage{url}            
\usepackage{booktabs}       
\usepackage{amsfonts}       
\usepackage{nicefrac}       
\usepackage{microtype}      
\usepackage{lipsum}
\usepackage{fancyhdr}       
\usepackage{graphicx}       
\graphicspath{{media/}}     
\usepackage{amssymb}
\usepackage{latexsym}
\usepackage{wrapfig}

\usepackage{graphicx}
\usepackage{mathtools}
\usepackage{amsfonts}
\usepackage{pifont}
\usepackage{booktabs}
\usepackage{todonotes}
\usepackage{siunitx}
\usepackage{microtype}
\usepackage{dirtytalk}
\usepackage{amsthm}
\usepackage{makecell} 
\usepackage{xcolor}
\usepackage[ruled,linesnumbered]{algorithm2e}
\usepackage{subcaption}
\usepackage{multirow}
\usepackage{placeins}

\SetKwInput{KwInput}{Input}
\SetKwInput{KwOutput}{Output}
\DeclarePairedDelimiterX{\inner}[2]{\langle}{\rangle}{#1, #2}

\newcounter{ArasCounter}

\newcounter{GittaCounter}

\newcounter{HolgerCounter}

\newcounter{PhilippCounter}

\newcommand{\revision}[1]{#1}

\newcommand*\widebar[1]{\@ifnextchar^{{\wide@bar{#1}{0}}}{\wide@bar{#1}{1}}}

\title{Learning-based adaption of robotic friction models
}

\author{
  Philipp Scholl \\
  Ludwig-Maximilians-Universität München \\
  Munich \\
  Germany\\
  \texttt{scholl@math.lmu.de} \\
   \And
  Maged Iskandar \\
  German Aerospace Center \\
  Weßling \\
  Germany\\
     \And
  Sebastian Wolf \\
  German Aerospace Center \\
  Weßling \\
  Germany\\
     \And
  Jinoh Lee \\
  German Aerospace Center \\
  Weßling \\
  Germany\\
     \And
    Aras Bacho \\
  Ludwig-Maximilians-Universität München \\
  Munich \\
  Germany\\
   \And
  Alexander Dietrich \\
  German Aerospace Center \\
  Weßling \\
  Germany\\
     \And
  Alin Albu-Schäffer \\
  German Aerospace Center \\
  Weßling \\
  Germany\\
     \And
  Gitta Kutyniok \\
  Ludwig-Maximilians-Universität München \\
  Munich Center for Machine Learning (MCML) \\
  Munich \\
  Germany\\
}

\begin{document}
\maketitle

\begin{abstract}
In the Fourth Industrial Revolution, wherein artificial intelligence and the automation of machines occupy a central role, the deployment of robots is indispensable. However, the manufacturing process using robots, especially in collaboration with humans, is highly intricate. In particular, modeling the friction torque in robotic joints is a longstanding problem due to the lack of a good mathematical description. This motivates the usage of data-driven methods in recent works. However, model-based and data-driven models often exhibit limitations in their ability to generalize beyond the specific dynamics they were trained on, as we demonstrate in this paper. To address this challenge, we introduce a novel approach based on residual learning, which aims to adapt an existing friction model to new dynamics using as little data as possible. We validate our approach by training a base neural network on a symmetric friction data set to learn an accurate relation between the velocity and the friction torque. Subsequently, to adapt to more complex asymmetric settings, we train a second network on a small dataset, focusing on predicting the residual of the initial network's output. By combining the output of both networks in a suitable manner, our proposed estimator outperforms the conventional model-based approach\revision{, an extended LuGre model,} and the base neural network significantly. Furthermore, we evaluate our method on trajectories involving external loads and still observe a substantial improvement, approximately 60-70\%, over the conventional approach. Our method does not rely on data with external load during training, eliminating the need for external torque sensors. This demonstrates the generalization capability of our approach, even with a small amount of data\textemdash{}\revision{less than a minute}\textemdash{}enabling adaptation to diverse scenarios based on prior knowledge about friction in different settings.\end{abstract}


\section{Introduction}

In the context of advancing human-robot collaboration in manufacturing \citep{ajoudani2018progress, kim2021human, ANGLERAUD2024102663, LIU2024102666}, where physical interaction between humans and robots is becoming increasingly important, the accurate estimation of interaction forces is critical for ensuring safety \citep{Aivaliotis2019, CHERUBINI20161, iskandar2023hybrid}. Achieving this often involves the precise estimation of external forces, which can be accomplished by mounting force-torque sensors on the robot's end-effector. Unfortunately, many industrial robots do not have these sensors due to their high cost \citep{Hägele2016}. This situation emphasizes the need for a reliable, cost-effective, and sensorless method for estimating external torque, which, in turn, necessitates the development of an accurate friction model.

A precise model of friction \citep{kermani2007friction} has the potential to enhance the functionality of robotic systems in multiple aspects. It plays a crucial role in achieving energy-efficient computations, enhancing the precision of dynamic simulations, improving control performance at the joint level \citep{dietrich2021practical,Xie2023,Ke2023,aivaliotis2023}, or detecting slip \citep{reinecke2014experimental}. To mitigate the undesirable effects of friction, a model-based friction compensation approach is commonly integrated into the control system. This approach finds widespread applications across multiple domains, including joint-level control \citep{iskandar2020,iskandar2022}, safe human-robot interaction \citep{de2008atlas,dietrich2019hierarchical,vogel2020ecosystem,iskandar2021}, and the enhancement of external torque estimation precision during interactions with the environment. \revision{Furthermore, friction is an important factor in the degradation curves of the components of a robot, thus, a precise friction estimation also improves the predictive maintenance \citep{AIVALIOTIS2021, AIVALIOTIS2021b}}. However, modeling the friction is highly complicated since it depends on a multitude of factors limiting the applicability of model-based approaches \citep{Rizos2005, Linderoth2013, khan2017review, hao2015observer, bittencourt2012static, bittencourt2013modeling}.

\subsection{Related work} \label{sec:related-work}

The success of neural networks (NN) in the last decade \citep{Krizhevsky2012, Silver2016, Fawzi2022, Jumper2021} inspired engineers to explore their application in modeling complex dynamics, including friction. \revision{An overview of data-driven approaches for friction modeling is given in Table~\ref{tab:related-work}.} \citet{Selmic2002}, for example, developed neural network models for friction that utilized specialized architectures with discontinuous activation functions. This design aimed to enhance the fitting of the observed friction data, ultimately reducing the number of required neurons, training time, and data. Another common approach is to use a hybrid method consisting of a simple parametric model and a complicated neural network. \citet{Ciliz2004, Ciliz2007} showed that this approach improves over both, a single parametric model and a single neural network, since it combines the flexibility of a neural network with the known dynamics captured by the simple parametric model, thus, introducing an inductive bias in the architecture.

\begin{table}[t]
    \caption{\revision{Overview of data-driven approaches to model the friction and their application.}}
    \label{tab:related-work}
    \footnotesize
    \centering
    \begin{tabular}{|c|c|c|}
        \hline
        Approach & Application goal & Method \\ \hline \hline 
        \cite{Selmic2002} & Friction estimation &  \makecell{Fully connected NN with \\ discontinuous activation functions}  \\ \hline
        \cite{Ciliz2004, Ciliz2007} & Motion control & \makecell{Hybrid of a parametric \\ model and NN} \\ \hline
        \cite{Huang2012} & Motion control & \makecell{Combination of \\ two NNs}  \\ \hline
        \cite{Guo2019} & Motion control & \makecell{Fully connected NN with \\ discontinuous activation functions } \\ \hline
        \cite{Hirose2017} & Friction estimation & \makecell{Long Short-Term Memory \\ (LSTM) network} \\ \hline
        \cite{tu2019modeling} & Friction estimation & \makecell{NN initialized using \\ Genetic Programming (GP)} \\ \hline
        \cite{Liu2021} & \makecell{External torque \\ estimation} & NN with sigmoid activation \\ \hline
    \end{tabular}
    
\end{table}

Recent research has aimed to leverage the structure of dynamics and learn multiple models to complement each other. \citet{Huang2012}, for instance, trained two neural networks specifically for modeling friction. \citet{Guo2019} took this a step further by learning additional, individual neural networks for the inertia matrix, the Coriolis torque, and the gravitational torque, which were subsequently combined into a single neural network. They also adapted discontinuous activation functions for friction torque. Most of the work, however, does not capture the hysteresis effect of friction, as they lack a dependence on the history. To model this, \citet{Hirose2017} applied a Long Short-Term Memory (LSTM) network, a type of recurrent neural network (RNN) that can naturally process sequences by maintaining a hidden state. Another approach was taken by \citet{tu2019modeling}, who used genetic algorithms (GA) to compute suitable initializations for the neural network weights. For human-robot interaction, it is crucial to be able to estimate the external torque of robot joints, which was the goal of \citet{Liu2021} by approximating the friction using neural networks.

\subsection{Our contributions}

The application of data-driven approaches to model friction in robotic joints is not a novel concept. However, previous approaches have exhibited limitations in their generality due to the absence of important components within the data. Specifically, these approaches lack the incorporation of different velocities, the reversal of directions, simultaneous movement of joints, and continuously varying loads. These conditions significantly influence the performance of data-driven approaches, as evidenced in Section~\ref{sec:prediction-of-the-base-models}. The results demonstrate that a network, that may outperform traditional model-based approaches on a dataset lacking some of these effects, ultimately fails when subjected to more complex data. Furthermore, dynamics also change due to wear, varying temperature and humidity, and other external factors. 

To address these challenges, we propose a novel strategy to adopt existing methods to unknown dynamics while requiring as little data as possible\textemdash{}only one movement in our case. To achieve this we train a neural network on the residual of a base estimator on new dynamics for which the base estimator fails, to be able to use the knowledge learned by the base estimator while improving its performance for the new dynamics. The key contributions of this paper are as follows:

\begin{itemize}
    \item[1.] \textbf{Adaption of a base model to new friction dynamics:} To tackle the challenge of adapting existing friction models to new dynamics with as little data as possible, we propose to learn the residuals using a neural network. 
    \item[2.] \textbf{Comprehensive evaluation:} We evaluate our approach using a data-driven base model of friction and compare it against a conventional model-based approach\revision{, an extended LuGre model}. We show that our approach then only requires data from \revision{ a single point-to-point movement that includes a velocity reversal} to be able to adapt the prediction of the friction torque and outperform the conventional approach and the base network for different velocities, while reversing the directions, moving the joints simultaneously, and continuously varying the external loads.
    \item[3.] \textbf{Integration into torque estimation framework:} To demonstrate the practical applicability of our adapted friction model, we integrate it into a torque estimation framework. By estimating the external torque applied to an object more precisely than the base network and the traditional approach, our approach enhances the overall precision and efficiency of robotic systems. We validate the accuracy of our torque estimation using external torque sensors, thereby ensuring the reliability of our proposed methodology. 
\end{itemize}

The successful adaption of existing friction models and their integration into the torque estimation framework holds promise for numerous applications in the field of robotics. This approach can potentially advance dynamic simulations, friction compensation techniques, and external torque estimation methodologies, enabling the development of more capable and adaptable robotic systems across various domains. By being able to adapt existing models to new dynamics, this approach leverages the wealth of existing knowledge and models, making it a powerful tool for addressing complex and dynamic scenarios in robotics.

\section{Methods}

The goal of this paper is to introduce a novel approach to adapt existing methods to new dynamics using as little data as possible. This is desirable for several reasons, including time and cost savings, prioritizing data quality over quantity, resource limitations, and reduced complexity. In Section~\ref{sec:model-baes-approach}, we start with the introduction of the conventional model-based approach we use as a baseline for comparison throughout our experiments. Afterward, in Section~\ref{sec:neural-network-based-approach}, we describe the neural network-based approach we will use as our base model. In Section~\ref{sec:adaption-to-new-dynamics}, we introduce our novel technique to adapt the neural network-based approach to unknown friction dynamics.

\revision{In the following, we describe the robot dynamics and our assumptions during the training of the different approaches to isolate the friction effect.} Consider the robot dynamics of the form 
\begin{equation}
M(q) \Ddot{q} + C(q,\Dot{q}) \Dot{q} + \tau_g(q) = \tau_m + \tau_f + \tau_{ext},
\end{equation}
where $q$ is the joint position vector, $\Dot{q}$ and $\Ddot{q} $ are the joint velocity and acceleration vectors, respectively, and $M(q)$ is the positive definite inertia matrix. The Coriolis and centrifugal matrix is denoted by $C(q,\Dot{q})$, the gravitational torque by $\tau_g(q)$, and the friction torque by $\tau_f$. Furthermore, the terms $\tau_m $ and  $\tau_{ext}$ describe the  motor joint torque and external joint torque, respectively.


\revision{To predict the external torque using the estimation of the friction torque, we rewrite the dynamics equation as}

\begin{equation}
\tau_{ext}= M(q) \Ddot{q} + C(q,\Dot{q}) \Dot{q} + \tau_g(q) - \tau_f - \tau_m.
\end{equation}

We assume full knowledge of the robot dynamic terms $(M(q),C(q,\Dot{q}),\tau_g(q))$, whereas $\tau_f$ is unknown. The motor torque in this case is directly measured as $\tau_m$.
If no load is applied ($\tau_{ext}=0$), the motor torque follows from the robot dynamics (assumed to be known) and the friction torque as
\begin{equation}
\tau_m = M(q) \Ddot{q} + C(q,\Dot{q}) \Dot{q} + \tau_g(q) - \tau_f .
\end{equation}

In the special case where we assume  \revision{constant, single-joint velocities or constant, low velocities} $(\Ddot{q}=0, C(q,\Dot{q}) \Dot{q}\approx0)$, the quadratic terms of individual joints can be ignored, while it is assured that the coupling terms are zero. As a result, the equation simplifies to 

\begin{equation} \label{eq:friction-low-constant-velocity}
\tau_m = \tau_g(q) - \tau_f.
\end{equation}

\revision{All the assumptions above will only be made during training to relate the motor torque to the friction effect as shown in Equation~\ref{eq:friction-low-constant-velocity}.}

\subsection{Model-based approach} \label{sec:model-baes-approach}

The friction torque $\tau_f$ can be mathematically described by incorporating several fundamental characteristics that define friction in the sliding regime, including static friction, Coulomb friction, viscous friction, and the Stribeck effect \citep{armstrong2012}. In the pure sliding regime, the static friction behavior $\tau_{\text{f,s}}$ can be represented by arbitrary functions, but a common model takes the form:
\begin{equation}\begin{split}\tau_{\text{f,s}}(\dot{q})&=g(\dot{q})+s(\dot{q})\\ g(\dot{q})&=sign(\dot{q})\left(F_{\text{c}}+(F_{\text{s}}- F_{\text{c}})e^{-\vert \dot{q}/ v_{\text{s}}\vert ^{\delta_{\text{s}}}}\right).\end{split}
\label{s_friction_model}
\end{equation}
The term $s(\dot{q})$ expresses the velocity-strengthening function that is well known as viscous friction. Typically, it is linearly proportional to the joint velocity $\dot{q}$ as $s(\dot{q})=F_v \dot{q}$, with the constant coefficient $F_v$. The velocity-strengthening friction effect could in the general case include a nonlinear form as shown in \citep{iskandar2019}. The function $g(\dot{q})$ describes the velocity-weakening behavior of the static friction. Also, $g(\dot{q})$ is alternatively called the Stribeck curve, because it captures the Stribeck effect, where $F_\text{c}$ is Coulomb friction, $F_\text{s}$ is static or stiction friction, $v_s$ is Stribeck velocity, and $\delta_{\text{s}}$ is the exponent parameter of the Stribeck-nonlinearity. 
The friction of the Harmonic-Drive (HD) gear-based robotic joint is known to be highly dependent on the temperature, which can be incorporated in the static and dynamic friction models \citep{wolf2018, iskandar2019}. Generally, the joint torque varies during the robot operation as it is configuration-dependent. This variation is reflected as a load effect in the joint friction torque, which can be included in Equation \eqref{s_friction_model} and results in
\begin{equation} \tau _{\mathrm{f},\mathrm{s}}(\dot {q}, \tau_\mathrm {l})=g(\dot {q}, \tau_\mathrm {l})+s(\dot{q},\tau_\mathrm {l})\,,
\end{equation}
where $\tau _{\mathrm{f},\mathrm{s}}(\dot {q}, \tau_\mathrm {l})$ denotes the static friction torque as a function of velocity $\dot{q}$ and load \revision{$\tau_\mathrm {l}$, which is mainly gravitational torque in our case,} for more details 
refer to \citep{iskandar2019}.
As the friction phenomenon by nature is nonlinear and continuous at zero velocity crossing, it is unpractical to use the static friction model which is discontinuous at velocity reversal. Therefore, the dynamic friction effect can be expressed as
\begin{equation} \tau _{\mathrm{f},\mathrm{d}}(\dot {q}, \tau_\mathrm {l})=f(z,\dot {q}, \tau_\mathrm {l})\,,\end{equation}
where $\tau _{\mathrm{f},\mathrm{d}}$ is the dynamic friction torque and $z$ is the internal friction state with its dynamics 
\begin{equation} \label{eq:dynamics-internal-friction-state}
    \frac{dz}{dt} = \dot{q} - \sigma_0 \frac{\vert{\dot{q}}\vert}{g(\dot {q}, \tau_\mathrm {l})} z.
\end{equation}

\noindent Equation~\eqref{eq:dynamics-internal-friction-state} can be rewritten as
\begin{equation}
\tau _{\mathrm{f},\mathrm{d}} = \sigma_0 z + \sigma_1 \dot{z} + s(\dot{q},\tau_\mathrm {l})\,.
\label{eq:dynamic_fric}
\end{equation}
The pre-sliding parameters are the bristle stiffness $\sigma_0$ and the micro-damping coefficient $\sigma_1$, for more details see \citep{Canudas1995,johanastrom2008revisiting}. 

\revision{Equation~\eqref{eq:dynamic_fric} represents a smooth and continuous expression that can describe the friction dynamically and extends the LuGre model \citep{Canudas1995} to incorporate load dependency, which is considered in this work as the conventional model-based approach.} While the bristles-based dynamic friction models show high accuracy in capturing the physical friction effects of the robotic joints, the difficulty of estimating and adapting their parameters remains. This limits the usage of such models in many scenarios and proves a strong motivation for data-driven approaches.

\subsection{Neural network based approach} \label{sec:neural-network-based-approach}
As our base model, we propose a data-driven model utilizing a neural network to learn the friction torque as a function of the gravity torque and its velocity. In our methodology, the focus lies on learning an unknown target function $f: X \rightarrow Y$ through the observation of input-output pairs $\lbrace(x_i, y_i)\rbrace_{i=1}^n \subset X \times Y$ with $f(x_i) \approx y_i$, where $X$ denotes the input space and $Y$ represents the output space. It is assumed, with some simplification, that the data points $x_i$ are sampled from a sequence of independently and identically distributed (iid) random variables, governed by a common probability density function $\mu$, which is supported on $X$, i.e. $x_i \sim \mu$ for all $i=1,\dots, n$ and $\mu(X)=1$. Furthermore, it is acknowledged that the observed output $y_i$ may potentially be corrupted by noise, thus introducing a degree of uncertainty. 

Our task now is to find a parameterized mapping $f_\theta: X \times \Theta \rightarrow Y$ (in our case, a neural network) that approximates the desired function $f$, where $\Theta$ represents a finite-dimensional parameter space. This is achieved by identifying a suitable cost function $C: X\times X\rightarrow \mathbb{R}$, which quantifies the discrepancy between the predicted outputs of $f_\theta$ and the true outputs, and minimizing the generalization error or risk, i.e., 
\begin{align*}
    \mathbb{E}_{X\sim \mu }[C(f,f_{\theta})]
\end{align*} over all $\theta\in \Theta$.
The generalization error $\mathbb{E}$ reflects the expected value of this cost function over the entire input space $X$, capturing the network's ability to generalize well to unseen data. By minimizing the generalization error, we strive to find the optimal set of parameters $\theta\in\Theta$ that minimizes the discrepancy between $f_\theta$ and the true function $f$, with the hope that $f \approx f_\theta$.

However, since, in general, we do not have access to the underlying probability measure $\mu$ and solely have a finite number of observations of input-output pairs $\lbrace(x_i, y_i)\rbrace_{i=1}^n \subset X \times Y$, we instead minimize the empirical risk or training error given by
\begin{align*}
    L(\theta)=\sum_{i=1}^n C(y_i,f_{\theta}(x_i))
\end{align*} over all $\theta\in \Theta$. An introduction to statistical learning theory can be found in \citet{Vapnik2000}.

\begin{figure}[h]
\centering
\includegraphics[width=8cm]{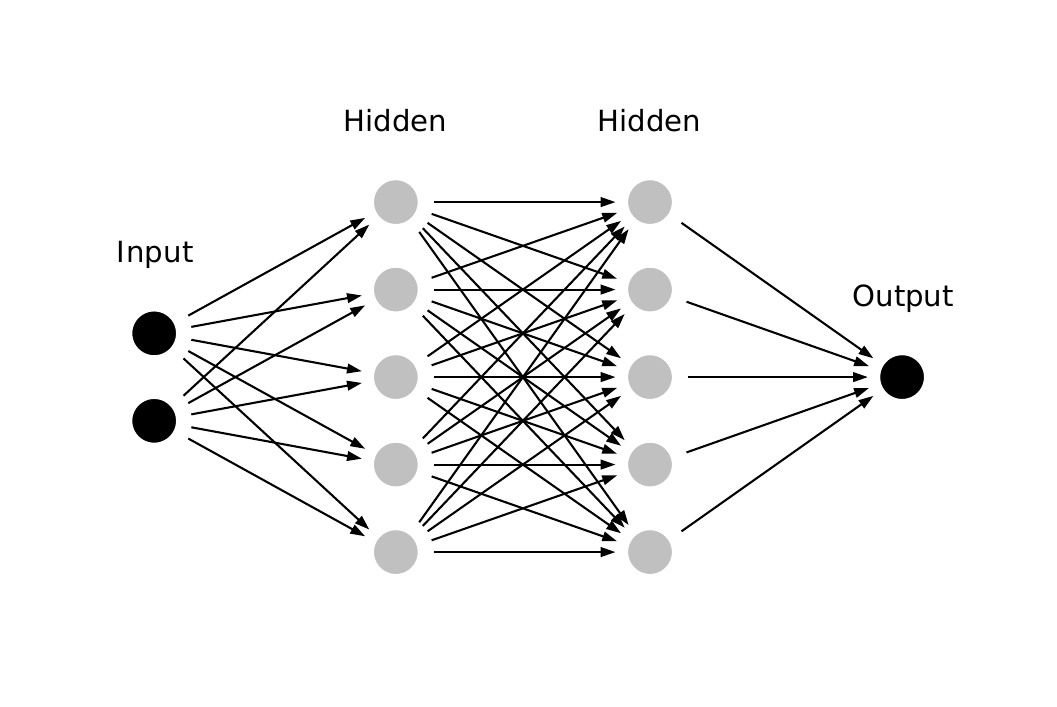}
\caption{The architecture of the base neural network: Fully connected neural network with 2 hidden layers, each consisting of 30 neurons (for space reasons, we only show five neurons here). }
\label{fig:neural-network-achitecture}
\end{figure}

In the framework of learning the torque friction, we choose a neural network as our parametric map denoted by $NN(\cdot, \theta)$, and the cost function is the squared difference of its inputs, i.e., the loss $L$ is the mean squared error.
A neural network is a function 
\begin{align*}
    NN=f_1\circ f_2\circ \dots \circ f_l
\end{align*} given by a composition of functions $f_i$ (the layers)
\begin{align}
    f_i(x;\theta_i) = \sigma(W_i x+b_i)
\end{align} where $\theta_i=(W_i,b_i)$ are the parameters,  $\sigma:\mathbb{R}\rightarrow \mathbb{R}$ is the so-called activation function, and $l\in \mathbb{N}$ is the number of layers of the neural network. Figure \ref{fig:neural-network-achitecture} shows the architecture of a two-hidden layer neural network which will be used for learning the torque friction. 

The idea in this paper is to train a neural network to model the friction torque by following \eqref{eq:friction-low-constant-velocity}. For this, we set $f_\theta=NN_{base}(\cdot;\theta): X\rightarrow Y$ with $X=\mathbb{R}^2$ and $Y=\mathbb{R}$, thereby allowing the neural network to depend on the gravitational torque $\tau_g(q(t_i))$ and the velocity $\dot q(t_i)$ at a given time step $t_i$.

As mentioned before, we then train the neural network by minimizing the mean squared error, i.e., the loss function is given by
\begin{equation} 
    L_{base}(\theta)=\sum_{i=0}^n\left(NN_{base}(\tau_g(q(t_i)),\dot{q}(t_i); \theta) - (\tau_m(t_i)-\tau_g(q(t_i)))\right)^2.
    \label{friction_and_g}
\end{equation}
$NN_{base}$ is a neural network with 2 hidden layers, each consisting of 30 neurons, as shown in Figure~\ref{fig:neural-network-achitecture}, which we will refer to as the base model in the following. As the activation function, we use the following exponential linear unit (ELU) function, given by,

\begin{equation}
    f(x)=
\begin{cases}
x, & x>0 \\
e^x-1, & x\leq0
\end{cases}
\end{equation}

\noindent which is a smoothed version of the rectified linear unit (ReLU) function.

\subsection{Adaption to new dynamics} \label{sec:adaption-to-new-dynamics}

Since most models of friction are best for the dynamics they are built and trained for, it is helpful if existing models can be easily extended to new situations without the need for big data sets to retrain the models completely. For that reason, we propose to build upon the existing models to not waste the previously acquired knowledge and train an additional neural network $NN_{add}$ on the residuals of the old model on the new data. Therefore, we minimize the following loss:

\begin{equation}
    \begin{split}
    L_{add}(\phi)= \sum_{i=0}^n & \left(NN_{add}(\tau_g^{add}(q(t_i)),sign(\dot{q}^{add}(t_i)); \phi) + 
    \tau_{f_{base}}(\tau_g^{add}(q(t_i)),\dot{q}^{add}(t_i))
    \right. \\ 
     & \left. - (\tau_m^{add}(t_i)-\tau_g^{add}(q(t_i)))\right)^2.        
    \end{split}
    \label{eq:Ladd}
\end{equation}

We allow the network $NN_{add}$ to depend on the gravity torque and the sign of the velocity; $\phi$ denotes the parameters of $NN_{add}$ we want to optimize and $\tau_{f_{base}}$ is the base model. In general, any base model works for $\tau_{f_{base}}$,  however, in this study, we utilize $NN_{base}$ as our base model, i.e., we set $\tau_{f_{base}}=NN_{base}$.

It is important to note that we only use the sign of the velocity as input for $NN_{add}$, rather than the full velocity. This choice is motivated by our objective to minimize the amount of data required for training. In our case, we will only use data from one movement with one velocity in both directions, which is insufficient to fully capture the complete  velocity dependence. Nevertheless, this is not a problem because the base model, $NN_{base}$, already accounts for the velocity dependence. Note further that using the sign of the velocity fits to the discontinuity of the friction at zero velocity. 

The new predictor for the small friction data set with varying load and directions is then

\begin{equation}
    \tau_{f_{pred}}(q, \dot{q})=\tau_{f_{base}}(q, \dot{q}; \theta^*) + NN_{add}(q, sign(\dot{q}); \phi^*), 
\end{equation}

\noindent where $\phi^*$ denotes the parameters of the additional model, found when minimizing $L_{add}(\phi)$. 

\section{Data sets}

\begin{figure}[h]
\centering
\includegraphics[width=7cm]{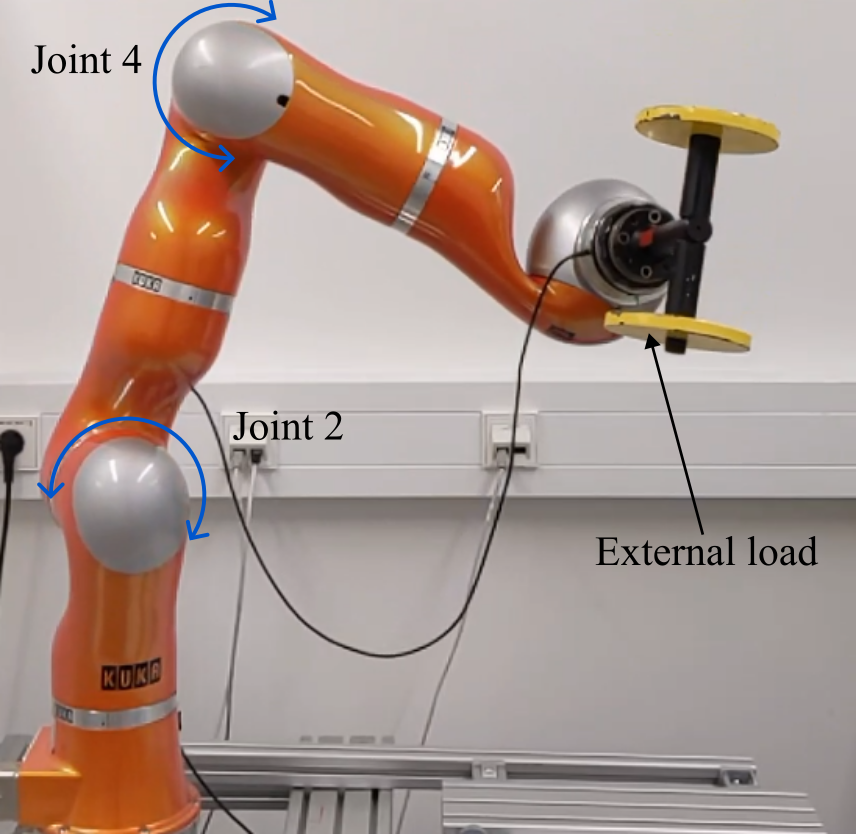}
\caption{The experimental setup: the robot is equipped with physical torque sensors in each joint output for the reference signals, while the measured motor current is used in the proposed method.}
\label{robot}
\end{figure}

Since the goal of this work is to adapt an existing approach to a set with different dynamics, we utilize two different data sets. Both data sets are collected to best capture the physical behavior of the friction torque. The experimental measurements are acquired from each robot joint separately at different constant velocities. In this study, the torque-controlled DLR-KUKA LWR-IV+ robot is used as a reference platform.
The full experimental setup is illustrated in Figure~\ref{robot}.   

This work aims to model the friction effects mainly resulting from the gear (HD), however, the used robot is also equipped with link-side torque sensors, which are used for validation only. The data is collected in two main subsets to best describe the friction effect and eliminate other unmodeled dynamics. 

\begin{figure}[h]
     \centering
     \begin{subfigure}[b]{0.45\textwidth}
         \centering
         \includegraphics[width=\textwidth]{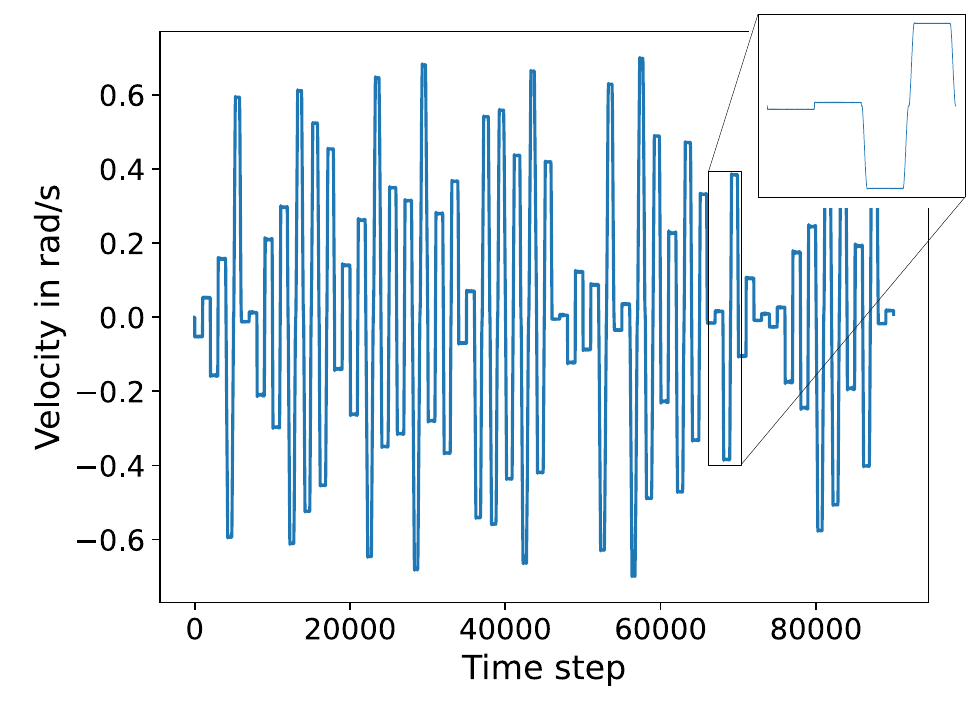}
         \caption{Velocity}
         \label{fig:Velocity-big-dataset}
     \end{subfigure}
     \hfill
     \begin{subfigure}[b]{0.49\textwidth}
         \centering
         \includegraphics[width=\textwidth]{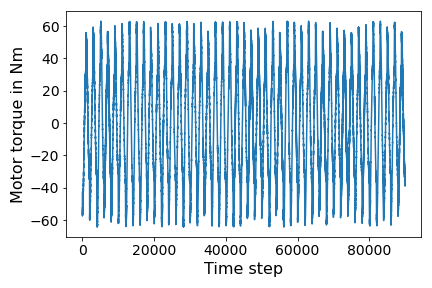}
         \caption{Motor torque}
         \label{fig:Motor-torque-big-dataset}
     \end{subfigure}
     \caption{Velocity and motor torque of the collected friction dataset for Joint 2.}
     \label{fig:Big-dataset-plots}
\end{figure}

\revision{For each data set we collect the position and velocity of the joints and the gravitational torque and applied motor current. The motor current $I_m$ is used to compute the motor torque $\tau_m=K_tI_m$, where $K_t$ is the motor torque constant taken from the manufacturer's data sheet. The gravitational torques in the robot joints are obtained through a model-based approach \revision{using recursive Newton-Euler formulation} which relies on the robot's mechanical design information, e.g., the mass and center of mass of each link\revision{, for which we rely on an accurate CAD model}.}

\revision{Notably, the load variation effect also appears in the no-load data as a result of gravitational torque in the robot joints. This can be seen from the measured data in Figure~\ref{fig:four-quadrants-collected-friction} and \ref{fig:four-quadrants-simultaneous-motion} in Section~\ref{sec:extended-dataset} as the friction torque varies even when the velocities are constant. The temperature effect is minimized by applying warm-up phases before each data collection routine. In this way, the temperature is assumed to be constant during operation.}

\subsection{Base data set}

\begin{figure}[t]
    \centering
     \begin{subfigure}[b]{0.45\textwidth}
         \centering
         \includegraphics[width=\textwidth]{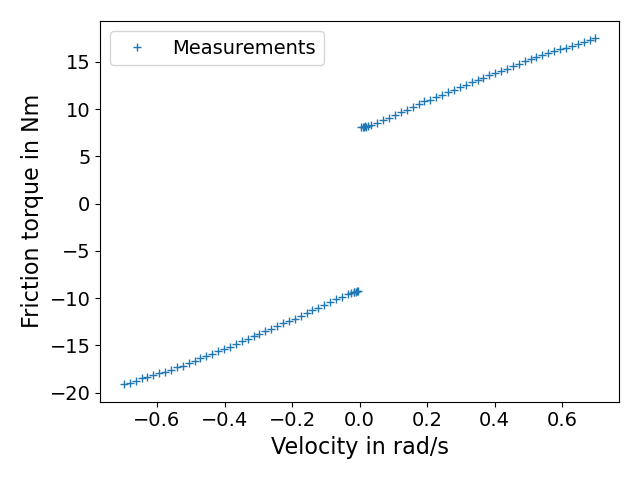}
         \caption{Joint 2}
         \label{fig:measurements-velocity-dependence-joint-2}
     \end{subfigure}
     \hfill
     \begin{subfigure}[b]{0.45\textwidth}
         \centering
         \includegraphics[width=\textwidth]{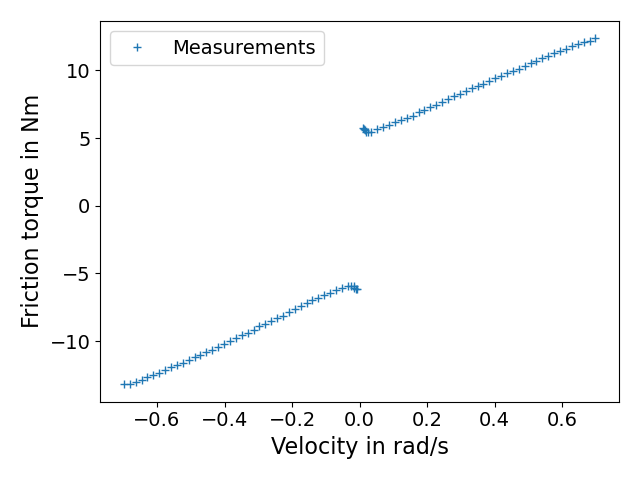}
         \caption{Joint 4}
         \label{fig:measurements-velocity-dependence-joint-4}
     \end{subfigure}
    \caption{The measured friction torque-velocity behavior for Joint 2 and Joint 4 of the DLR-KUKA LWR-IV+ robot for the base data set.}
    \label{fig:measurements-velocity-dependence}
\end{figure}

The base data set aims to capture the static friction behavior; therefore, the robot joints were excited to follow \revision{single-joint} constant velocities. \revision{Thus, \eqref{eq:friction-low-constant-velocity} holds and can be used to compute the friction.} We will refer to this data set as the collected friction data set. Intensive experimental measurements have been carried out to cover the entire operational velocity and position ranges of the individual joints; the velocity and motor torque of Joint 2 can be seen in Figure~\ref{fig:Big-dataset-plots}. In this way, the static friction is characterized in a fine resolution and separated from other effects as shown in Figure~\ref{fig:measurements-velocity-dependence}.

\subsection{Extended data set} \label{sec:extended-dataset}

\begin{figure}[h]
     \centering
     \begin{subfigure}[b]{0.49\textwidth}
         \centering
         \includegraphics[width=\textwidth]{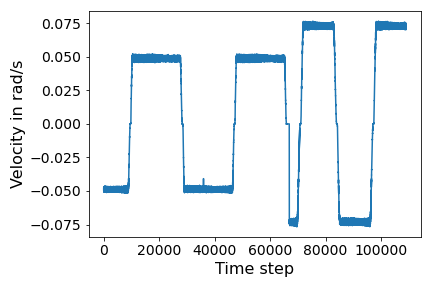}
         \caption{Velocity}
         \label{fig:Velocity-small-dataset}
     \end{subfigure}
     \hfill
     \begin{subfigure}[b]{0.49\textwidth}
         \centering
         \includegraphics[width=\textwidth]{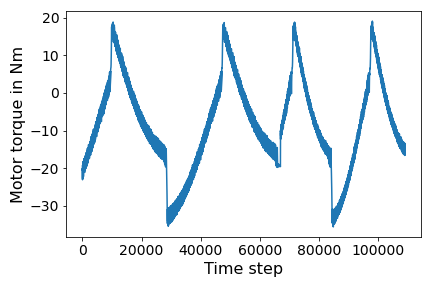}
         \caption{Motor torque}
         \label{fig:Motor-torque-small-dataset}
     \end{subfigure}
     \caption{Velocity and motor torque of the small dataset with varying directions and without external load for Joint 2.}
     \label{fig:small-dataset-plots}
\end{figure}

The extended data set is collected dynamically by applying different velocities sequentially while different external loads are attached to the robot end-effector and the joints are moved simultaneously. We refer to this data set as the small data set with varying loads and directions. The velocity and motor torque for this dataset without external torque are displayed in Figure~\ref{fig:small-dataset-plots}. Additionally, the motor current $I_m$ that corresponds to each joint motion is collected, which is linearly proportional to the motor torque as $\tau_m = K_t I_m$, where $K_t$ is the motor torque constant. Ideally, at a constant velocity and known gravity torque (see Equation 4), the motor torque purely corresponds to the friction torque. The gravitational torques in the robot joints are obtained through a model-based approach which relies on the robot's mechanical design information, e.g., the mass and center of mass of each link. This assumes that the mass and center of mass of the robot links are perfectly known. The load variation effect also appears in the no-load data as a result of gravitational torque in the robot joints. This can be seen from the measured data in Figure~\ref{fig:NN-no-load} in Section~\ref{sec:prediction-of-the-base-models} as the friction torque varies even when the velocities are constant. The temperature effect is minimized by applying warm-up phases before each data collection routine. In this way, the temperature is assumed to be constant during operation.

\begin{figure}[t]
     \centering
     \begin{subfigure}[b]{0.35\textwidth}
        \includegraphics[width=\textwidth]{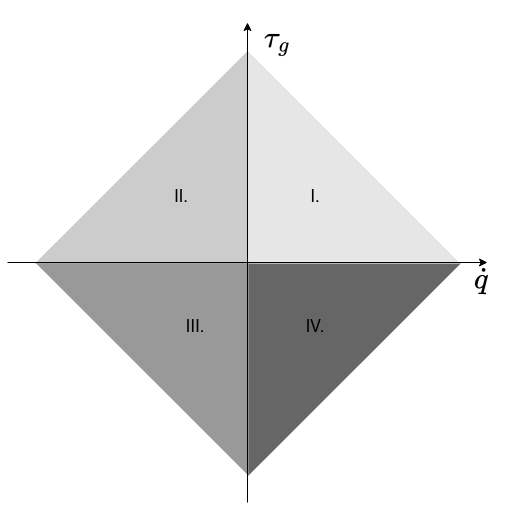}
        \caption{The four quadrants of friction force.}  
        \label{fig:four-quadrants-of-friction-force}
     \end{subfigure} 
     \hfill

     \begin{subfigure}[b]{0.49\textwidth}
         \centering
    \includegraphics[width=\textwidth]{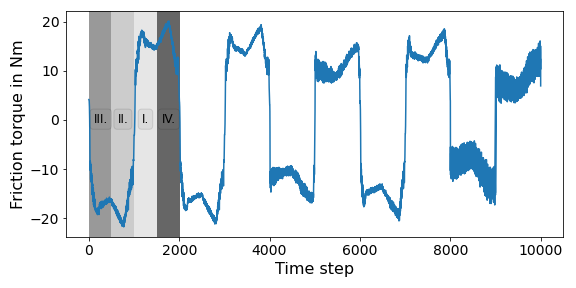}
         \caption{Collected friction data set.}  \label{fig:four-quadrants-collected-friction}
     \end{subfigure}
     \hfill
    \begin{subfigure}[b]{0.49\textwidth}
         \centering
         \includegraphics[width=\textwidth]{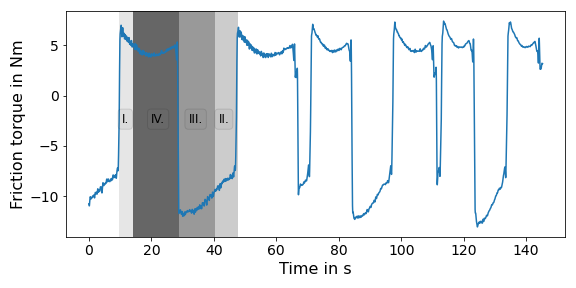}
         \caption{Simultaneous motion data set.} \label{fig:four-quadrants-simultaneous-motion}
     \end{subfigure}
    \centering
    \caption{Four quadrants of friction force: The different grey tones show the different quadrants the friction is going through. The friction stays in each quadrant for the exact same time in the collected friction data set, but the times vary strongly for the simultaneous motion data set.}
    \label{fig:four-quadrants-of-friction-force-both-data-sets}
\end{figure}

The behavior of friction torque in robotic joints varies across the four quadrants, as defined in Figure~\ref{fig:four-quadrants-of-friction-force}, due to changes in the interaction between velocity and gravitational torque \citep{Abayebas2021}: 
In the first and third quadrants, they have the same sign, while they have different signs in the second and fourth. 
The difference between the two datasets investigated in this paper can be visualized by examining the quadrants during one movement from each dataset, as shown in Figure~\ref{fig:four-quadrants-of-friction-force-both-data-sets}. It is evident that the base dataset features entirely symmetric quadrants, whereas the additional dataset demonstrates a highly asymmetric trajectory across the quadrants.

\section{Experimental results}

In this section, we present the results of the numerical experiments. In Subsection~\ref{sec:prediction-of-the-base-models}, we show the results of $NN_{base}$ on both the symmetric and asymmetric datasets and compare the performance to the conventional approach. 
These experiments reveal that the neural network-based approach performs similarly to the conventional approach and both struggle with the asymmetric data set. Consequently, in Subsection~\ref{sec:prediciton-of-the-adapted-models}, we demonstrate the improvements achieved by adapting $NN_{base}$ using $NN_{add}$.

\subsection{Prediction of the base model} \label{sec:prediction-of-the-base-models}

\begin{figure}[h]
     \centering
     \begin{subfigure}[b]{0.49\textwidth}
         \centering
         \includegraphics[width=\textwidth]{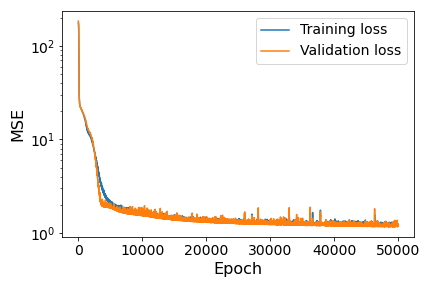}
         \caption{Joint 2}
         \label{fig:NN-loss-joint-2}
     \end{subfigure}
     \hfill
     \begin{subfigure}[b]{0.49\textwidth}
         \centering
         \includegraphics[width=\textwidth]{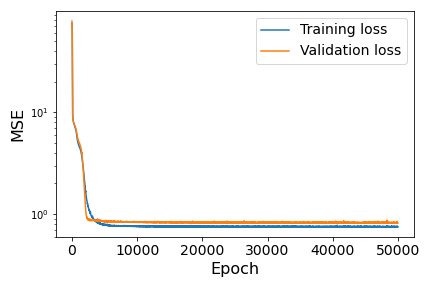}
         \caption{Joint 4}
         \label{fig:NN-loss-joint-4}
     \end{subfigure}
     \caption{The training and validation loss $L_{base}$ for Joint 2 and Joint 4 for $NN_{base}$.}
     \label{fig:NN-loss}
\end{figure}

\begin{figure}[h]
     \centering
     \begin{subfigure}[b]{0.49\textwidth}
         \centering
         \includegraphics[width=\textwidth]{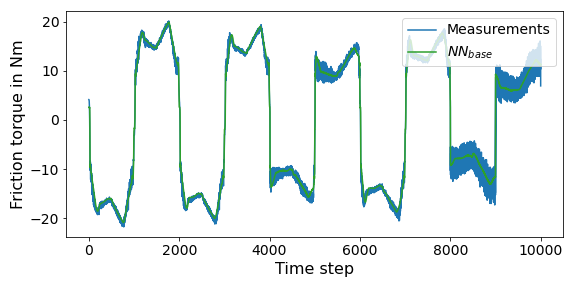}
         \caption{Friction Joint 2}
         \label{fig:NN-no-load-joint-2}
     \end{subfigure}
     \hfill
     \begin{subfigure}[b]{0.49\textwidth}
         \centering
         \includegraphics[width=\textwidth]{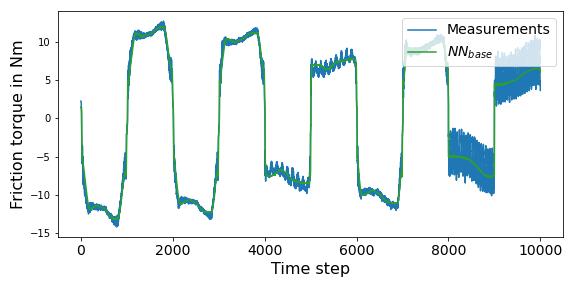}
         \caption{Friction Joint 4}
         \label{fig:NN-no-load-joint-4}
     \end{subfigure}
    \begin{subfigure}[b]{0.49\textwidth}
         \centering
         \includegraphics[width=\textwidth]{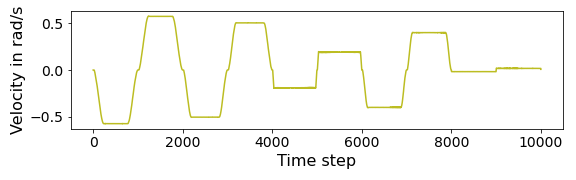}
         \caption{Velocity Joint 2}
         \label{fig:velocity-no-load-joint-2}
     \end{subfigure}
     \hfill
     \begin{subfigure}[b]{0.49\textwidth}
         \centering
         \includegraphics[width=\textwidth]{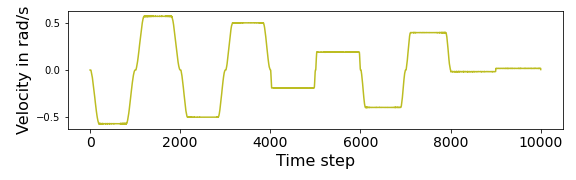}
         \caption{Velocity Joint 4}
         \label{fig:velocity-no-load-joint-4}
     \end{subfigure}
     \caption{Figures (a) and (b) show the true friction torque of the collected friction data for Joints 2 and 4, respectively, which was withheld during the training and validation phase, and the estimate of the neural network. The mean absolute error was $0.79 Nm$ and $0.60 Nm$, respectively. Figures (c) and (d) show the velocity for Joints 2 and 4, respectively.}
     \label{fig:NN-no-load}
\end{figure}

The base network $NN_{base}$ is trained on a subset of the collected friction data set using the Adam optimizer \citep{Kingma2015Ba} with a learning rate of 0.01 for 50,000 training steps.
Since for each velocity, the joint moves exactly once in both directions, the data contains more measurements in the low-velocity regime than in the high-velocity regime. 
As a consequence, the loss function tends to prioritize the low-velocity region, resulting in high precision for low velocities and low precision for high velocities. To address this,  we downsample the data, ensuring an equal number of data points for each velocity, as depicted in Figure~\ref{fig:Big-dataset-plots}. Besides balancing the precision across velocity ranges, this downsampling also significantly improves runtime, particularly in the low-velocity regime where many data points are very similar.

\begin{figure}[h]
     \centering
     \begin{subfigure}[b]{0.45\textwidth}
         \centering
         \includegraphics[width=\textwidth]{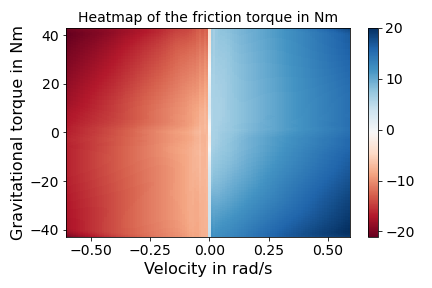}
         \caption{Heatmap Joint 2}
         \label{fig:NN-velocity-position-dependence-heatmap-joint-2}
     \end{subfigure}
     \hfill
    \begin{subfigure}[b]{0.45\textwidth}
         \centering
         \includegraphics[width=\textwidth]{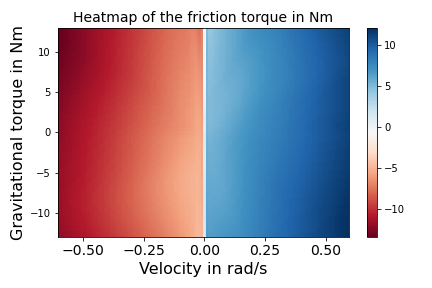}
         \caption{Heatmap Joint 4}
         \label{fig:NN-velocity-position-dependence-heatmap-joint-4}
     \end{subfigure}
     \hfill
     \begin{subfigure}[b]{0.48\textwidth}
         \centering
         \includegraphics[width=\textwidth]{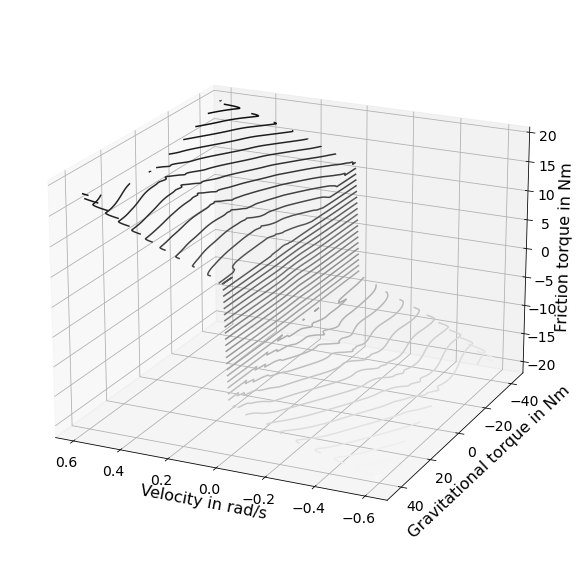}
         \caption{3-dimensional plot Joint 2}
         \label{fig:NN-velocity-position-dependence-3d-plot-joint-2}
     \end{subfigure}
      \hfill
     \begin{subfigure}[b]{0.48\textwidth}
         \centering
         \includegraphics[width=\textwidth]{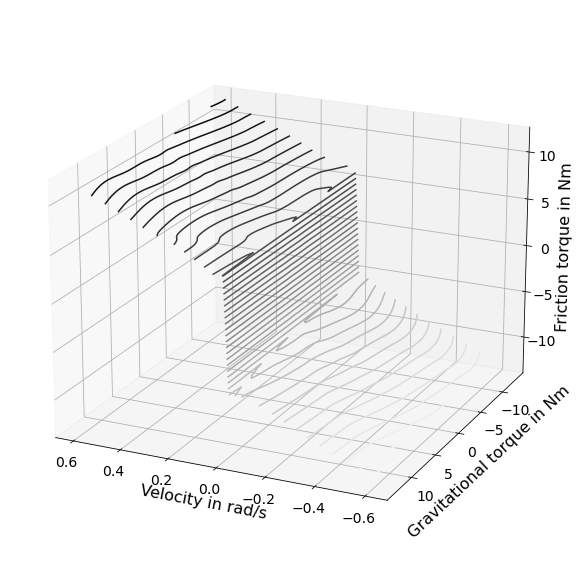}
         \caption{3-dimensional plot Joint 4}
         \label{fig:NN-velocity-position-dependence-3d-plot-joint-4}
     \end{subfigure}
     \caption{Dependence of the friction on the velocity and the position as captured by the base model $NN_{base}$ for Joint 2 and 4.}
     \label{fig:NN-velocity-position-dependence}
\end{figure}
\begin{figure}[h]
    \centering
     \begin{subfigure}[b]{0.49\textwidth}
         \centering
         \includegraphics[width=\textwidth]{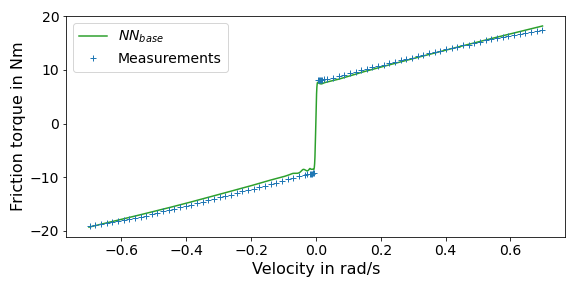}
         \caption{Joint 2}
         \label{fig:NN-velocity-dependence-joint-2}
     \end{subfigure}
     \hfill
     \begin{subfigure}[b]{0.49\textwidth}
         \centering
         \includegraphics[width=\textwidth]{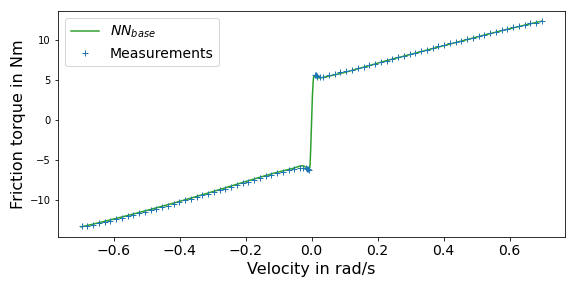}
         \caption{Joint 4}
         \label{fig:NN-velocity-dependence-joint-4}
     \end{subfigure}
    \caption{Dependence of the friction on the velocity as modeled by the base model $NN_{base}$.}
    \label{fig:NN-velocity-dependence}
\end{figure}

Figure~\ref{fig:NN-loss} displays the behavior of the loss function $L_{base}$ on both the training and validation data during the training of $NN_{base}$ for Joint 2 and 4, illustrating that the model effectively converges robustly to a local minimum and can be expected to generalize well, since the validation loss is close to the training loss. 
To assess the performance of the base neural network $NN_{base}$ trained on the collected friction data set, Figure~\ref{fig:NN-no-load} presents the comparison between its prediction and the measurements. The network demonstrates its capability to accurately approximate the friction torque, showcasing its effectiveness.
In the next step, we explore whether the data-driven model $NN_{base}$ learned a meaningful relationship between velocity and friction. To investigate this, we evaluate the model on a grid spanning from $-0.7$ to $0.7$~$rad/s$ for the velocity and from $-43$ to $43$~$Nm$ for the gravitational torque for Joint~2 ($-13$ to $13$~$Nm$ for Joint~4). Figure~\ref{fig:NN-velocity-position-dependence} visualizes this evaluation as both a heatmap and a 3D plot, while Figure~\ref{fig:NN-velocity-dependence} presents the averaged results over the gravitational torque. These plots clearly demonstrate that the base neural network has learned a physically meaningful friction model, suggesting strong potential for generalization, as further supported by the results in Figure~\ref{fig:NN-no-load}.

\begin{figure}[h]
     \centering
     \begin{subfigure}[b]{0.45\textwidth}
         \centering
    \includegraphics[width=\textwidth]{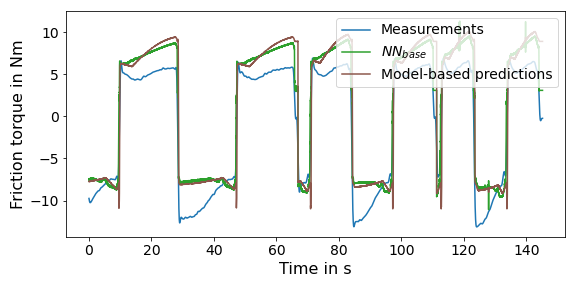}
         \caption{Friction Joint 2}
         \label{fig:NN-discs-asym-joint-2}
     \end{subfigure}
     \hfill
    \begin{subfigure}[b]{0.45\textwidth}
         \centering
         \includegraphics[width=\textwidth]{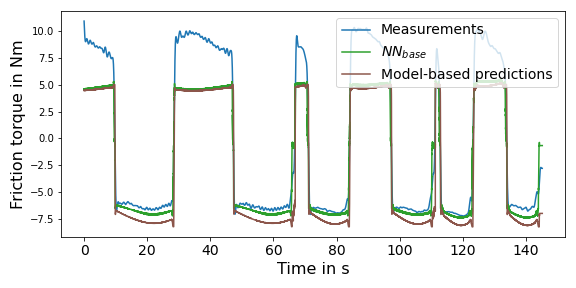}
         \caption{Friction Joint 4}
         \label{fig:NN-discs-asym-joint-4}
     \end{subfigure}
     \hfill
    \begin{subfigure}[b]{0.45\textwidth}
         \centering
         \includegraphics[width=\textwidth]{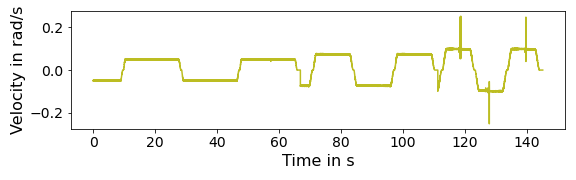}
         \caption{Velocity Joint 2}
         \label{fig:velocity-discs-asym-joint-2}
     \end{subfigure}
     \hfill
     \begin{subfigure}[b]{0.45\textwidth}
         \centering
         \includegraphics[width=\textwidth]{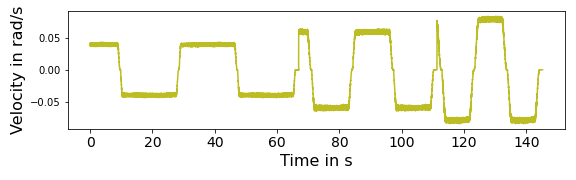}
         \caption{Velocity Joint 4}
         \label{fig:velocity-discs-asym-joint-4}
     \end{subfigure}
    \hfill
    \begin{subfigure}[b]{0.45\textwidth}
         \centering
         \includegraphics[width=\textwidth]{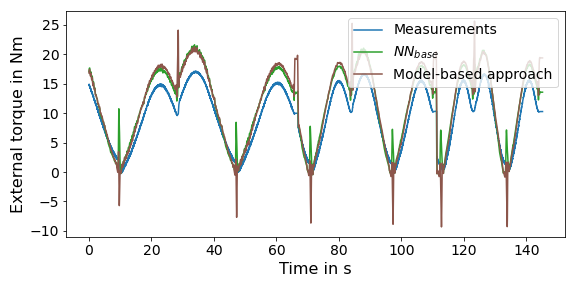}
         \caption{External torque Joint 2}
         \label{fig:NN-discs-asym-external-torque-joint-2}
     \end{subfigure}
     \hfill
     \begin{subfigure}[b]{0.45\textwidth}
         \centering
         \includegraphics[width=\textwidth]{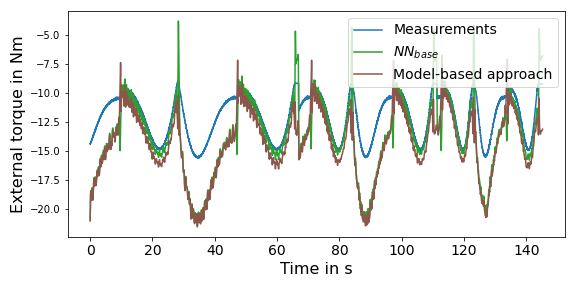}
         \caption{External torque Joint 4}
         \label{fig:NN-discs-asym-external-torque-joint-4}
     \end{subfigure}
    \centering
    \caption{Performance of the neural network versus the model-based approach on the small data set with asymmetric load. In (a) and (b) the friction is estimated for Joint 2 and 4. On Joint 2, the neural network achieved an average error of $2.40 Nm$, while the conventional approach had an average error of $2.68 Nm$. On Joint 4, the neural network achieved an average error of $1.91 Nm$, while the conventional approach had an average error of $2.45 Nm$. (c) and (d) show the velocities of the respective joints. In (e) and (f) the friction estimates are used to estimate the external torque, which is shown after denoising.}
    \label{fig:NN-discs-asym}
\end{figure}

\begin{figure}[h]
     \centering
     \begin{subfigure}[b]{0.45\textwidth}
         \centering
    \includegraphics[width=\textwidth]{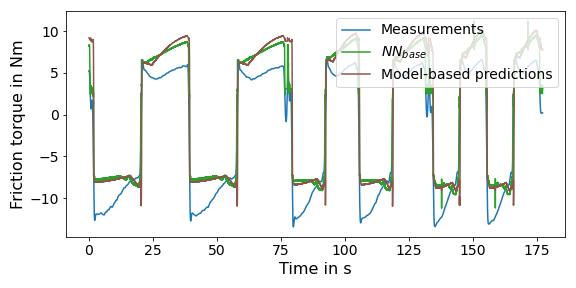}
         \caption{Friction Joint 2}
         \label{fig:NN-discs-sym-joint-2}
     \end{subfigure}
     \hfill
    \begin{subfigure}[b]{0.45\textwidth}
         \centering
         \includegraphics[width=\textwidth]{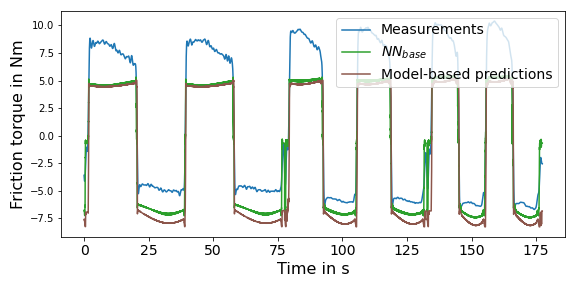}
         \caption{Friction Joint 4}
         \label{fig:NN-discs-sym-joint-4}
     \end{subfigure}
     \hfill
    \begin{subfigure}[b]{0.45\textwidth}
         \centering
         \includegraphics[width=\textwidth]{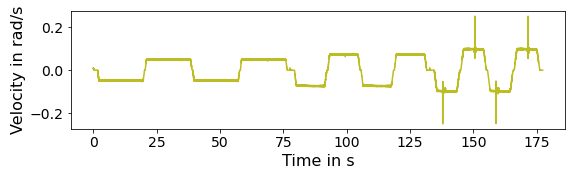}
         \caption{Velocity Joint 2}
         \label{fig:NN-velocity-discs-sym-joint-2}
     \end{subfigure}
     \hfill
     \begin{subfigure}[b]{0.45\textwidth}
         \centering
         \includegraphics[width=\textwidth]{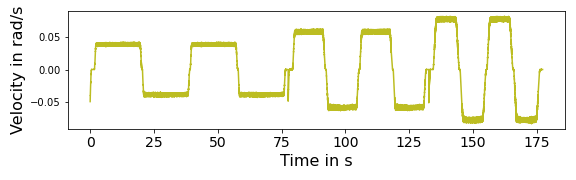}
         \caption{Velocity Joint 4}
         \label{fig:NN-velocity-discs-sym-joint-4}
     \end{subfigure}
          \hfill
    \begin{subfigure}[b]{0.45\textwidth}
         \centering
         \includegraphics[width=\textwidth]{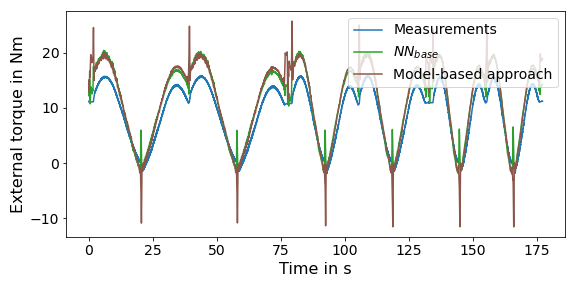}
         \caption{External torque Joint 2}
         \label{fig:NN-discs-sym-external-torque-joint-2}
     \end{subfigure}
     \hfill
     \begin{subfigure}[b]{0.45\textwidth}
         \centering
         \includegraphics[width=\textwidth]{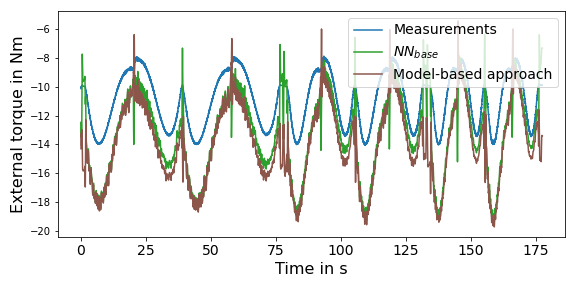}
         \caption{External torque Joint 4}
         \label{fig:NN-discs-sym-external-torque-joint-4}
     \end{subfigure}
    \centering
    \caption{Performance of the base model $NN_{base}$ versus the model-based approach on the small data set with symmetric load. In (a) and (b) the friction is estimated for Joint~2 and 4. On Joint~2, the base model achieved an average error of $2.51 Nm$, while the conventional approach had an average error of $2.89 Nm$. On Joint~4 the base model achieved an average error of $2.48 Nm$, while the conventional approach had an average error of $3.1 Nm$. (c) and (d) show the velocities of the respective joints. In (e) and (f) the friction estimates are used to estimate the external torque, which is shown after denoising.}
    \label{fig:NN-discs-sym}
\end{figure}

As the neural network effectively captured the dynamics of friction for the base data set, the next step is to assess its performance on the extended data set. Figures~\ref{fig:NN-discs-asym} and \ref{fig:NN-discs-sym} reveal that the base network $NN_{base}$ outperforms the conventional method, but it performs significantly worse than it did for the collected friction data set. In Figures~\ref{fig:NN-discs-asym} (e) and (f), as well as Figures~\ref{fig:NN-discs-sym} (e) and (f), the  friction estimates are utilized to predict the external torque, illustrating the prediction error introduced through the error in friction modeling. 

\FloatBarrier

\subsection{Prediction of the adapted model} \label{sec:prediciton-of-the-adapted-models} 

\begin{figure}[h]
     \centering
     \begin{subfigure}[b]{0.49\textwidth}
         \centering
         \includegraphics[width=\textwidth]{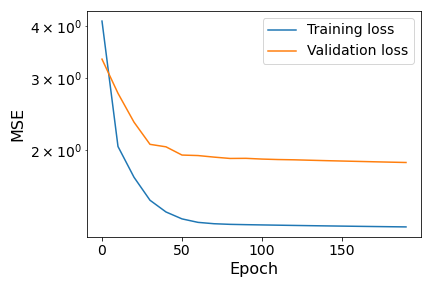}
         \caption{Joint 2}
         \label{fig:signed-posnet-loss-joint-2}
     \end{subfigure}
     \hfill
     \begin{subfigure}[b]{0.49\textwidth}
         \centering
         \includegraphics[width=\textwidth]{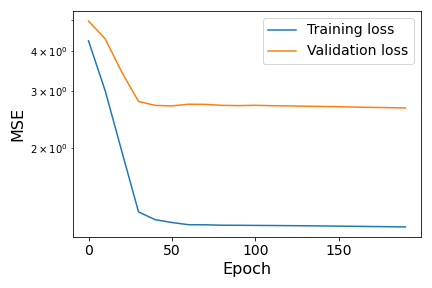}
         \caption{Joint 4}
         \label{fig:signed-posnet-loss-joint-4}
     \end{subfigure}
     \caption{The training and validation loss $L_{add}$ for Joints 2 and 4 for $NN_{add}$.}
     \label{fig:signed-posnet-loss}
\end{figure}

To improve the prediction of $NN_{base}$ on the extended data set, we train the additive network $NN_{add}$ as described in Section~\ref{sec:adaption-to-new-dynamics} by minimizing $L_{add}$ in Equation~\eqref{eq:Ladd}. Specifically, we provide only one velocity from the small data set without external load as training data. This does not lead to overfitting, since $NN_{add}$ depends solely on the gravitational torque and the sign of the velocity. The benefit of using only data without external load during training is that this does not require any external torque sensor. The architecture of the additional model $NN_{add}$ mirrors that of the base model, however, it only uses one hidden layer. The training process is also the same, but $NN_{add}$ is trained for only 200 epochs. Figure~\ref{fig:signed-posnet-loss} displays the training and validation loss during training, showcasing robust minimization across epochs, but a slight difference in training and validation loss. 

\begin{figure}[h]
     \centering
     \begin{subfigure}[b]{0.49\textwidth}
         \centering
    \includegraphics[width=\textwidth]{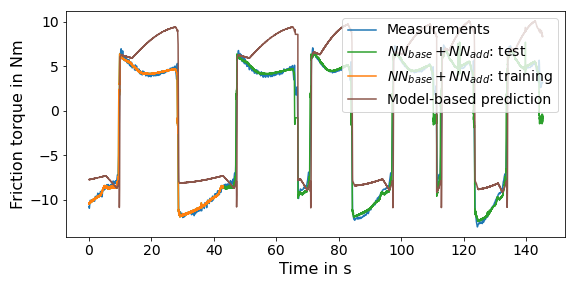}
         \caption{Friction Joint 2}
         \label{fig:signed-posnet-friction-joint-2}
     \end{subfigure}
     \hfill
    \begin{subfigure}[b]{0.49\textwidth}
         \centering
         \includegraphics[width=\textwidth]{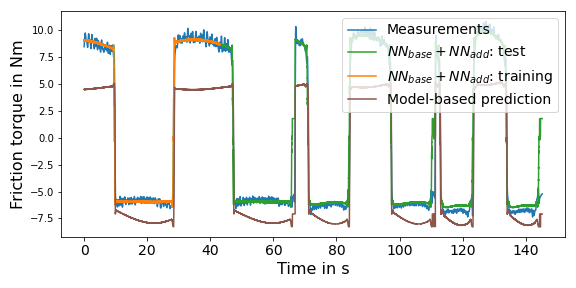}
         \caption{Friction Joint 4}
         \label{fig:signed-posnet-friction-joint-4}
     \end{subfigure}
     \hfill
    \begin{subfigure}[b]{0.49\textwidth}
         \centering
         \includegraphics[width=\textwidth]{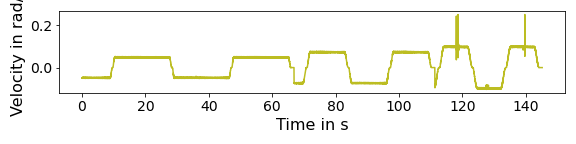}
         \caption{Velocity Joint 2}
     \end{subfigure}
     \hfill
     \begin{subfigure}[b]{0.49\textwidth}
         \centering
         \includegraphics[width=\textwidth]{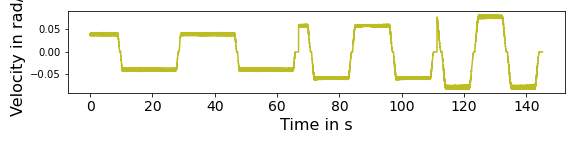}
         \caption{Velocity Joint 4}
     \end{subfigure}
          \hfill
    \centering
    \caption{Performance of the combined network $NN_{base}+NN_{add}$ versus the model-based approach on the small data set without external load. In (a) and (b) the friction is estimated for Joint 2 and 4. On Joint 2, the combined neural network achieved an average error (on the test data) of $0.48 Nm$, while the conventional approach had an average error of $2.62 Nm$. On Joint 4, the neural network achieved an average error of $ 0.66 Nm$, while the conventional approach had an average error of $2.33 Nm$. (c) and (d) show the velocities of the respective joints.}
    \label{fig:signed-posnet-no-load-small}
\end{figure}

\begin{figure}[h]
    \centering
     \begin{subfigure}[b]{0.49\textwidth}
         \centering
         \includegraphics[width=\textwidth]{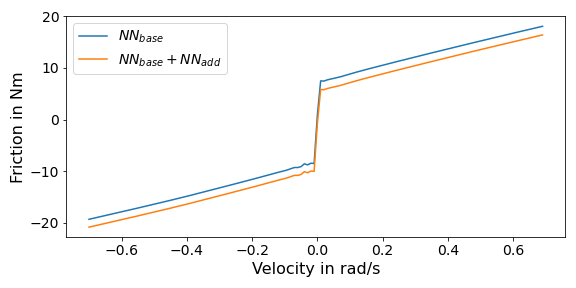}
         \caption{Joint 2}
         \label{fig:signed-posnet-velocity-dependence-joint-2}
     \end{subfigure}
     \hfill
     \begin{subfigure}[b]{0.49\textwidth}
         \centering
         \includegraphics[width=\textwidth]{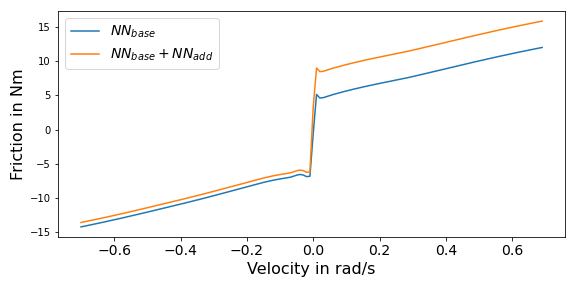}
         \caption{Joint 4}
         \label{fig:signed-posnet-velocity-dependence-joint-4}
     \end{subfigure}
    \caption{Dependence of the friction on the velocity as modelled by the combined neural network $NN_{base}+NN_{add}$ versus as modelled by the base neural network.}
    \label{fig:signed-PosNet-velocity-dependence}
\end{figure}

\begin{figure}[h]
     \centering
     \begin{subfigure}[b]{0.45\textwidth}
         \centering
         \includegraphics[width=\textwidth]{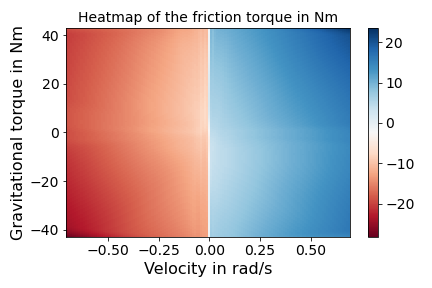}
         \caption{Heatmap Joint 2}
         \label{fig:signed PosNet-velocity-position-dependence-heatmap-joint-2}
     \end{subfigure}
     \hfill
    \begin{subfigure}[b]{0.45\textwidth}
         \centering
         \includegraphics[width=\textwidth]{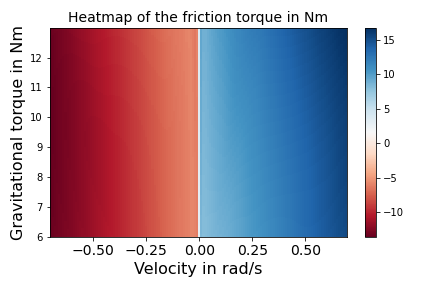}
         \caption{Heatmap Joint 4}
         \label{fig:signed PosNet-velocity-position-dependence-heatmap-joint-4}
     \end{subfigure}
     \hfill
     \begin{subfigure}[b]{0.49\textwidth}
         \centering
         \includegraphics[width=\textwidth]{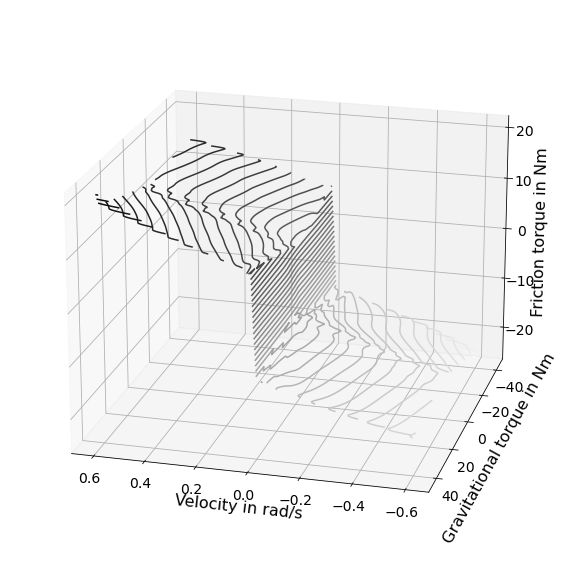}
         \caption{3 dimensional plot Joint 2}
         \label{fig:signed PosNet-velocity-position-dependence-3d-plot-joint-2}
     \end{subfigure}
      \hfill
     \begin{subfigure}[b]{0.49\textwidth}
         \centering
         \includegraphics[width=\textwidth]{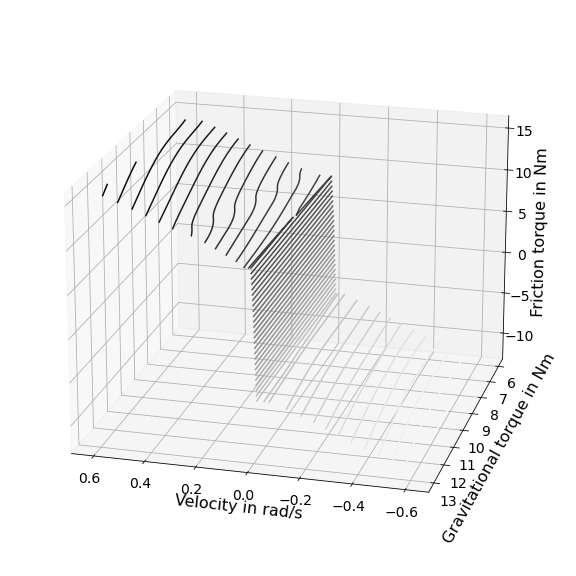}
         \caption{3 dimensional plot Joint 4}
         \label{fig:signed PosNet-velocity-position-dependence-3d-plot-joint-4}
     \end{subfigure}
     \caption{Dependence of the friction on the velocity and the position as captured by the combined neural network $NN_{base}+NN_{add}$ for Joint 2 and 4.}
     \label{fig:signed-PosNet-velocity-position-dependence}
\end{figure}

Figure~\ref{fig:signed-posnet-no-load-small} demonstrates the accuracy of the friction modeling achieved by the adapted approach, as described in Section~\ref{sec:adaption-to-new-dynamics}, in comparison to $NN_{base}$ and the conventional approach, even though only one velocity was observed during the additional training period (43s). 
Furthermore, Figures~\ref{fig:signed-PosNet-velocity-dependence} and~\ref{fig:signed-PosNet-velocity-position-dependence} reveal that this extension of the original neural network has a minimal impact on the velocity dependence, particularly preserving the underlying relationship that would be lost through the retraining of any network components.

\begin{figure}[h]
     \centering
     \begin{subfigure}[b]{0.45\textwidth}
         \centering
    \includegraphics[width=\textwidth]{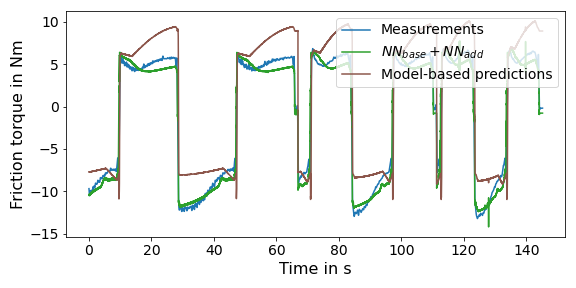}
         \caption{Friction Joint 2}
         \label{fig:signed PosNet-discs-asym-joint-2}
     \end{subfigure}
     \hfill
    \begin{subfigure}[b]{0.45\textwidth}
         \centering
         \includegraphics[width=\textwidth]{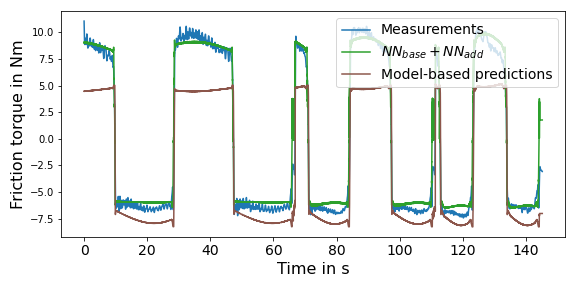}
         \caption{Friction Joint 4}
         \label{fig:signed PosNet-discs-asym-joint-4}
     \end{subfigure}
     \hfill
    \begin{subfigure}[b]{0.45\textwidth}
         \centering
         \includegraphics[width=\textwidth]{Images/NN/Joint_2/Velocity_discs_asym_joint_2.png}
         \caption{Velocity Joint 2}
         \label{fig:signed-PosNet-velocity-discs-asym-joint-2}
     \end{subfigure}
     \hfill
     \begin{subfigure}[b]{0.45\textwidth}
         \centering
         \includegraphics[width=\textwidth]{Images/NN/Joint_4/Velocity_discs_asym_joint_4.png}
         \caption{Velocity Joint 4}
         \label{fig:signed-PosNet-velocity-discs-asym-joint-4}
     \end{subfigure}
    \hfill
    \begin{subfigure}[b]{0.45\textwidth}
         \centering
         \includegraphics[width=\textwidth]{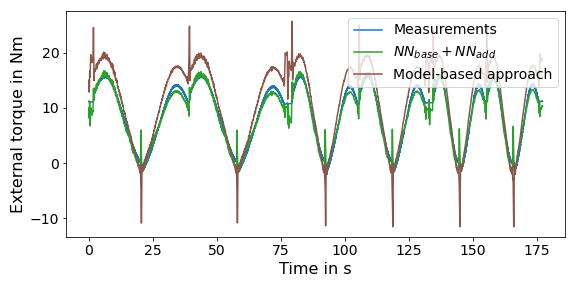}
         \caption{External torque Joint 2}
         \label{fig:signed-PosNet-discs-asym-external-torque-joint-2}
     \end{subfigure}
     \hfill
     \begin{subfigure}[b]{0.45\textwidth}
         \centering
         \includegraphics[width=\textwidth]{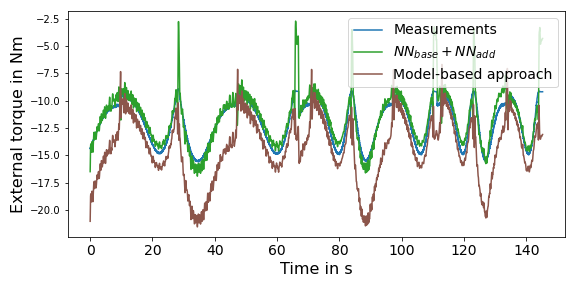}
         \caption{External torque Joint 4}
         \label{fig:signed-PosNet-discs-asym-external-torque-joint-4}
     \end{subfigure}
    \centering
    \caption{Performance of the combined neural network $NN_{base}+NN_{add}$ versus the model-based approach on the asymmetric load data. In (a) and (b) the friction is estimated for Joint 2 and 4. On Joint 2, the combined neural network achieved an average error of $0.87 Nm$, while the conventional approach had an average error of $2.66 Nm$. On Joint 4, the neural network achieved an average error of $0.70 Nm$, while the conventional approach had an average error of $2.45 Nm$. (c) and (d) show the velocities of the respective joints. In (e) and (f) the friction estimates are used to estimate the external torque, which is shown after denoising.}
    \label{fig:signed-PosNet-discs-asym}
\end{figure}

\begin{figure}[h]
     \centering
     \begin{subfigure}[b]{0.45\textwidth}
         \centering
    \includegraphics[width=\textwidth]{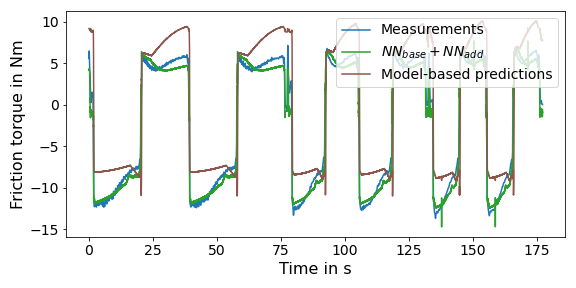}
         \caption{Friction Joint 2}
         \label{fig:signed PosNet-discs-sym-joint-2}
     \end{subfigure}
     \hfill
    \begin{subfigure}[b]{0.45\textwidth}
         \centering
         \includegraphics[width=\textwidth]{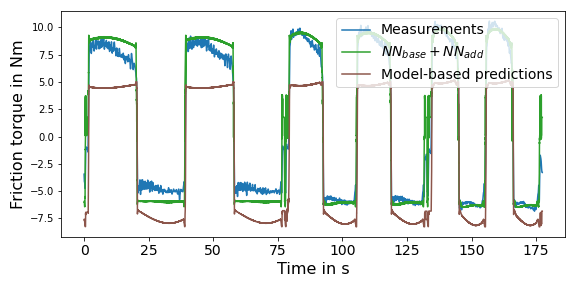}
         \caption{Friction Joint 4}
         \label{fig:signed PosNet-discs-sym-joint-4}
     \end{subfigure}
     \hfill
    \begin{subfigure}[b]{0.45\textwidth}
         \centering
         \includegraphics[width=\textwidth]{Images/NN/Joint_2/Velocity_discs_sym_joint_2.png}
         \caption{Velocity Joint 2}
         \label{fig:signed-PosNet-velocity-discs-sym-joint-2}
     \end{subfigure}
     \hfill
     \begin{subfigure}[b]{0.45\textwidth}
         \centering
         \includegraphics[width=\textwidth]{Images/NN/Joint_4/Velocity_discs_sym_joint_4.png}
         \caption{Velocity Joint 4}
         \label{fig:signed-PosNet-velocity-discs-sym-joint-4}
     \end{subfigure}
    \hfill
    \begin{subfigure}[b]{0.45\textwidth}
         \centering
         \includegraphics[width=\textwidth]{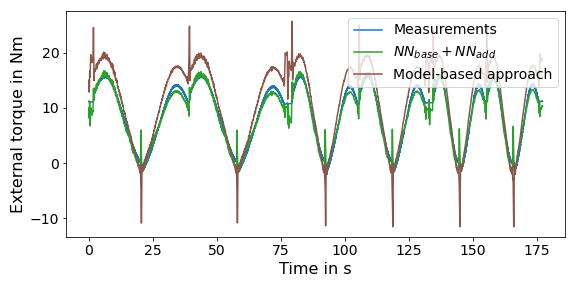}
         \caption{External torque Joint 2}
         \label{fig:signed-PosNet-discs-sym-external-torque-joint-2}
     \end{subfigure}
     \hfill
     \begin{subfigure}[b]{0.45\textwidth}
         \centering
         \includegraphics[width=\textwidth]{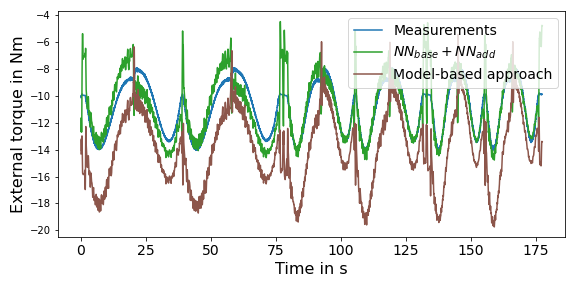}
         \caption{External torque Joint 4}
         \label{fig:signed-PosNet-discs-sym-external-torque-joint-4}
     \end{subfigure}
    \centering
    \caption{Performance of the combined neural network $NN_{base}+NN_{add}$ versus the model-based approach on the symmetric load data. In (a) and (b) the friction is estimated for Joint 2 and 4. On Joint 2, the combined neural network achieved an average error of $0.80 Nm$, while the conventional approach had an average error of $2.88 Nm$. On Joint 4, the neural network achieved an average error of $0.86 Nm$, while the conventional approach had an average error of $3.09 Nm$. (c) and (d) show the velocities of the respective joints. In (e) and (f) the friction estimates are used to estimate the external torque, which is shown after denoising.}
    \label{fig:signed-PosNet-discs-sym}
\end{figure}

The combined approach not only exhibits a significant improvement for the small data set with varying directions when no external load is applied but also when asymmetric and symmetric external loads are present, as demonstrated in Figure~\ref{fig:signed-PosNet-discs-asym} (a) and (b) and Figure~\ref{fig:signed-PosNet-discs-sym} (a) and (b), respectively. By leveraging the accurate prediction of the adapted method for the friction torque we can estimate the external torque via the robot dynamics. Contrary to the prediction using only $NN_{base}$ in Figure~\ref{fig:NN-discs-asym} (e) and (f) and Figure~\ref{fig:NN-discs-sym} (e) and (f), this estimate strongly outperforms the conventional approach, see Figure~\ref{fig:signed-PosNet-discs-asym} (e) and (f) and Figure~\ref{fig:signed-PosNet-discs-sym} (e) and (f) for the asymmetric and symmetric load cases, respectively.

\FloatBarrier

\subsection{Comparisons with other data-driven methods}

\revision{To the best of our knowledge, this is the first work focusing on the adaptation of existing friction estimation models. Therefore, we primarily compared the base method with the adapted method, which is method-agnostic and can be applied to other methods presented in the literature. To ensure clarity, we concentrated in Sections \ref{sec:prediction-of-the-base-models} and \ref{sec:prediciton-of-the-adapted-models} on classical methods (vanilla neural network and LuGre model), highlighting the benefits of residual learning.}

\revision{However, a comparative analysis with other data-driven approaches, as outlined in the literature review in Section~\ref{sec:related-work}, is of significant interest. To accomplish this, we re-implemented the methods proposed by \citet{Selmic2002} and by \citet{Ciliz2004, Ciliz2007}.}

\revision{\cite{Selmic2002} introduced a fully connected neural network with discontinuous activation functions to model the discontinuity of friction at zero velocity. For this purpose they applied the standard Sigmoid functions $\sigma(x)=\frac{1}{1+e^{-x}}$ and additionally
the Sigmoid jump approximation functions}
\begin{equation}
    \phi_k(x)=\begin{cases}
        0, &  \text{ for } x<0,\\
        (1-e^{-x})^k, &  \text{ for } x\geq0.
    \end{cases}
\end{equation}
\revision{An extensive hyperparameter search, similar to the one performed for $NN_{base}$, showed that the network works best with 1 hidden layer comprising 30 neurons. Among these neurons, 20 used the standard Sigmoid as the activation function, while the remaining neurons applied the Sigmoid jump approximation function, with $k$ varying from 1 to 10 for each neuron, as proposed by Selmic and Lewis \cite{Selmic2002}. The network was optimized using ADAM with a learning rate of 0.01 over 50,000 steps.}

\revision{\citet{Ciliz2004, Ciliz2007} combine a neural network with a parametric approach to incorporate the flexibility of data-driven approaches with common knowledge about friction by modelling the friction torque $\tau_f$ as}
\begin{equation} \label{eq:ciliz-tomizuka}
    \hat{\tau}_f(\dot q, \tau_q) = NN(\dot q, \tau_q)+\beta_10.5(1+sign(\dot q) + \beta_20.5(1-sign(\dot q)
\end{equation}
\revision{named \emph{neural network with adaptive coulomb friction (NNACM)}. Here, $\beta_1$ and $\beta_2$ denote learnable parameters.We extended the original NNACM model to allow the neural network to depend on the gravitational torque, as shown in \eqref{eq:ciliz-tomizuka}, to ensure a fair comparison with the other approaches presented in this work. Through an extensive hyperparameter search, we determined that using one hidden layer with 50 neurons yielded optimal results. The network was optimized using ADAM with a learning rate of 0.01 over 50,000 steps.}

\revision{The results of both methods on the extended data set with asymmetric and symmetric loads can be observed in Table~\ref{tab:comparison-related-work}. It is evident from the table that both approaches perform slightly better than the extended LuGre model and the base neural network. However, the combined neural network proposed to learn the new dynamics still outperformed them by a margin, suggesting that these approaches also struggled with the new dynamics.}

\begin{table}[h]
    \caption{\revision{Comparisons between the base methods from Section~\ref{sec:model-baes-approach} and \ref{sec:neural-network-based-approach} and the adapted approach from Section~\ref{sec:adaption-to-new-dynamics} with the approaches from \citet{Selmic2002} and \citet{Ciliz2004, Ciliz2007} on joint 2.}}
    \label{tab:comparison-related-work}
    \centering
    \begin{tabular}{|c|c|c|}
        \hline
        Method & \makecell{Error with the \\asymmetric load } & \makecell{Error with the \\symmetric load } \\ \hline \hline
        \makecell{Conventional model \\ (ext. LuGre)}  & 2.68Nm & 2.89Nm \\ \hline
        \makecell{$NN_{base}$} & 2.40Nm & 2.51Nm\\ \hline
        \makecell{\cite{Selmic2002}} & 2.17Nm & 2.37Nm \\ \hline
        \makecell{\cite{Ciliz2004} \\ \cite{Ciliz2007}} & 2.33Nm & 2.42Nm \\ \hline
        \makecell{$\mathbf{NN_{base}+NN_{add}}$} & \textbf{0.87Nm} & \textbf{0.80Nm} \\ \hline
    \end{tabular}
\end{table}

\FloatBarrier

\section{Summary and Conclusion}

Due to the lack of a precise mathematical description of friction torque in robotic joints, model-based approaches struggle to capture its behavior accurately. This, in turn, hinders robot control in various situations, especially when dealing with new movements or environments. To address this challenge, we proposed an approach that adapts existing friction models to new dynamics using a minimal amount of data, which enhances efficiency, prioritizes data quality over quantity, and addresses resource limitations.

Our method leverages neural networks to learn the residuals of a base model on new dynamics, significantly improving model accuracy with as little as 43 seconds of additional data. Importantly, our approach does not require specific domain knowledge and eliminates the need for external torque sensors, reducing the overall cost of robotic systems. Although this paper primarily demonstrates the use of a neural network as the base model, Equation \eqref{eq:Ladd} permits the adaptation of any base model, thus enabling the incorporation of specialized data-driven or model-based techniques. Our method has been thoroughly tested with highly detailed data encompassing diverse velocities, loads, and movement directions, and its predictions have been validated using torque sensors.

However, neural networks' black-box nature and occasional difficulties in generalizing to out-of-distribution data remain challenges. Addressing these issues while retaining the advantages of data-driven friction torque learning will be important areas for future research. Besides this, we consider the incorporation of the adaption approach proposed in this work into an online learning procedure as a valuable future direction.

\section*{Acknowledgements}
P. Scholl and G. Kutyniok acknowledge support by LMUexcellent, funded by the Federal Ministry of Education and Research (BMBF) and the Free State of Bavaria under the Excellence Strategy of the Federal Government and the Länder as well as by the Hightech Agenda Bavaria. 

Furthermore, G. Kutyniok was supported in part by the DAAD programme Konrad Zuse Schools of Excellence in Artificial Intelligence, sponsored by the Federal Ministry of Education and Research. 
G. Kutyniok also acknowledges support from the Munich Center for Machine Learning (MCML) as well as the German Research Foundation under Grants DFG-SPP-2298, KU 1446/31-1 and KU 1446/32-1 and under
Grant DFG-SFB/TR 109 and Project C09.

M. Iskandar, A. Dietrich, and J. Lee were supported by MOTIE/KEIT under Grants 20014485 and 37520014398.

\bibliographystyle{abbrvnat}
\bibliography{references}  

\providecommand{\noopsort}[1]{}\providecommand{\singleletter}[1]{#1}%
\begin{thebibliography}{48}
\providecommand{\natexlab}[1]{#1}
\providecommand{\url}[1]{\texttt{#1}}
\expandafter\ifx\csname urlstyle\endcsname\relax
  \providecommand{\doi}[1]{doi: #1}\else
  \providecommand{\doi}{doi: \begingroup \urlstyle{rm}\Url}\fi

\bibitem[Abayebas et~al.(2021)Abayebas, Yumbla, Yi, Lee, Park, and Moon]{Abayebas2021}
M.~Abayebas, F.~Yumbla, J.-S. Yi, W.~Lee, J.~Park, and H.~Moon.
\newblock Passivity guaranteed dynamic friction model with temperature and load correction: Modeling and compensation for collaborative industrial robot.
\newblock \emph{IEEE Access}, PP:\penalty0 1--1, 04 2021.
\newblock \doi{10.1109/ACCESS.2021.3076308}.

\bibitem[Aivaliotis et~al.(2019)Aivaliotis, Aivaliotis, Gkournelos, Kokkalis, Michalos, and Makris]{Aivaliotis2019}
P.~Aivaliotis, S.~Aivaliotis, C.~Gkournelos, K.~Kokkalis, G.~Michalos, and S.~Makris.
\newblock Power and force limiting on industrial robots for human-robot collaboration.
\newblock \emph{Robotics and Computer-Integrated Manufacturing}, 59:\penalty0 346--360, 2019.
\newblock ISSN 0736-5845.
\newblock \doi{https://doi.org/10.1016/j.rcim.2019.05.001}.

\bibitem[Aivaliotis et~al.(2021{\natexlab{a}})Aivaliotis, Arkouli, Georgoulias, and Makris]{AIVALIOTIS2021}
P.~Aivaliotis, Z.~Arkouli, K.~Georgoulias, and S.~Makris.
\newblock Degradation curves integration in physics-based models: Towards the predictive maintenance of industrial robots.
\newblock \emph{Robotics and Computer-Integrated Manufacturing}, 71:\penalty0 102177, 2021{\natexlab{a}}.
\newblock ISSN 0736-5845.
\newblock \doi{https://doi.org/10.1016/j.rcim.2021.102177}.

\bibitem[Aivaliotis et~al.(2021{\natexlab{b}})Aivaliotis, Arkouli, Kaliakatsos-Georgopoulos, and Makris]{AIVALIOTIS2021b}
P.~Aivaliotis, Z.~Arkouli, D.~Kaliakatsos-Georgopoulos, and S.~Makris.
\newblock Prediction assessment methodology for maintenance applications in manufacturing.
\newblock \emph{Procedia CIRP}, 104:\penalty0 1494--1499, 2021{\natexlab{b}}.
\newblock ISSN 2212-8271.
\newblock \doi{https://doi.org/10.1016/j.procir.2021.11.252}.
\newblock 54th CIRP CMS 2021 - Towards Digitalized Manufacturing 4.0.

\bibitem[Ajoudani et~al.(2018)Ajoudani, Zanchettin, Ivaldi, Albu-Sch{\"a}ffer, Kosuge, and Khatib]{ajoudani2018progress}
A.~Ajoudani, A.~M. Zanchettin, S.~Ivaldi, A.~Albu-Sch{\"a}ffer, K.~Kosuge, and O.~Khatib.
\newblock Progress and prospects of the human--robot collaboration.
\newblock \emph{Autonomous Robots}, 42:\penalty0 957--975, 2018.

\bibitem[Angleraud et~al.(2024)Angleraud, Ekrekli, Samarawickrama, Sharma, and Pieters]{ANGLERAUD2024102663}
A.~Angleraud, A.~Ekrekli, K.~Samarawickrama, G.~Sharma, and R.~Pieters.
\newblock Sensor-based human–robot collaboration for industrial tasks.
\newblock \emph{Robotics and Computer-Integrated Manufacturing}, 86:\penalty0 102663, 2024.
\newblock ISSN 0736-5845.
\newblock \doi{https://doi.org/10.1016/j.rcim.2023.102663}.

\bibitem[Armstrong-Helouvry(2012)]{armstrong2012}
B.~Armstrong-Helouvry.
\newblock \emph{Control of machines with friction}, volume 128.
\newblock Springer Science \& Business Media, 2012.

\bibitem[Bittencourt and Axelsson(2013)]{bittencourt2013modeling}
A.~C. Bittencourt and P.~Axelsson.
\newblock Modeling and experiment design for identification of wear in a robot joint under load and temperature uncertainties based on friction data.
\newblock \emph{IEEE/ASME transactions on mechatronics}, 19\penalty0 (5):\penalty0 1694--1706, 2013.

\bibitem[Bittencourt and Gunnarsson(2012)]{bittencourt2012static}
A.~C. Bittencourt and S.~Gunnarsson.
\newblock Static friction in a robot joint—modeling and identification of load and temperature effects.
\newblock 2012.

\bibitem[Canudas~de Wit et~al.(1995)Canudas~de Wit, Olsson, Astrom, and Lischinsky]{Canudas1995}
C.~Canudas~de Wit, H.~Olsson, K.~Astrom, and P.~Lischinsky.
\newblock A new model for control of systems with friction.
\newblock \emph{IEEE Transactions on Automatic Control}, 40\penalty0 (3):\penalty0 419--425, 1995.
\newblock \doi{10.1109/9.376053}.

\bibitem[Cherubini et~al.(2016)Cherubini, Passama, Crosnier, Lasnier, and Fraisse]{CHERUBINI20161}
A.~Cherubini, R.~Passama, A.~Crosnier, A.~Lasnier, and P.~Fraisse.
\newblock Collaborative manufacturing with physical human–robot interaction.
\newblock \emph{Robotics and Computer-Integrated Manufacturing}, 40:\penalty0 1--13, 2016.
\newblock ISSN 0736-5845.
\newblock \doi{https://doi.org/10.1016/j.rcim.2015.12.007}.

\bibitem[Ciliz and Tomizuka(2004)]{Ciliz2004}
M.~Ciliz and M.~Tomizuka.
\newblock Neural network based friction compensation in motion control.
\newblock \emph{Electronics Letters}, 40:\penalty0 752 -- 753, 07 2004.
\newblock \doi{10.1049/el:20040500}.

\bibitem[De~Santis et~al.(2008)De~Santis, Siciliano, De~Luca, and Bicchi]{de2008atlas}
A.~De~Santis, B.~Siciliano, A.~De~Luca, and A.~Bicchi.
\newblock An atlas of physical human--robot interaction.
\newblock \emph{Mechanism and Machine Theory}, 43\penalty0 (3):\penalty0 253--270, 2008.

\bibitem[Dietrich and Ott(2019)]{dietrich2019hierarchical}
A.~Dietrich and C.~Ott.
\newblock Hierarchical impedance-based tracking control of kinematically redundant robots.
\newblock \emph{IEEE Transactions on Robotics}, 36\penalty0 (1):\penalty0 204--221, 2019.

\bibitem[Dietrich et~al.(2021)Dietrich, Wu, Bussmann, Harder, Iskandar, Englsberger, Ott, and Albu-Sch{\"a}ffer]{dietrich2021practical}
A.~Dietrich, X.~Wu, K.~Bussmann, M.~Harder, M.~Iskandar, J.~Englsberger, C.~Ott, and A.~Albu-Sch{\"a}ffer.
\newblock Practical consequences of inertia shaping for interaction and tracking in robot control.
\newblock \emph{Control Engineering Practice}, 114:\penalty0 104875, 2021.

\bibitem[Fawzi et~al.(2022)Fawzi, Balog, Huang, Hubert, Romera-Paredes, Barekatain, Novikov, Ruiz, Schrittwieser, Swirszcz, Silver, Hassabis, and Kohli]{Fawzi2022}
A.~Fawzi, M.~Balog, A.~Huang, T.~Hubert, B.~Romera-Paredes, M.~Barekatain, A.~Novikov, F.~Ruiz, J.~Schrittwieser, G.~Swirszcz, D.~Silver, D.~Hassabis, and P.~Kohli.
\newblock Discovering faster matrix multiplication algorithms with reinforcement learning.
\newblock \emph{Nature}, 610:\penalty0 47--53, 10 2022.
\newblock \doi{10.1038/s41586-022-05172-4}.

\bibitem[Guo et~al.(2019)Guo, Pan, and Yu]{Guo2019}
K.~Guo, Y.~Pan, and H.~Yu.
\newblock Composite learning robot control with friction compensation: A neural network-based approach.
\newblock \emph{IEEE Transactions on Industrial Electronics}, 66\penalty0 (10):\penalty0 7841--7851, 2019.
\newblock \doi{10.1109/TIE.2018.2886763}.

\bibitem[H{\"a}gele et~al.(2016)H{\"a}gele, Nilsson, Pires, and Bischoff]{Hägele2016}
M.~H{\"a}gele, K.~Nilsson, J.~N. Pires, and R.~Bischoff.
\newblock \emph{Industrial Robotics}, pages 1385--1422.
\newblock Springer International Publishing, Cham, 2016.
\newblock ISBN 978-3-319-32552-1.
\newblock \doi{https://doi.org/10.1007/978-3-319-32552-1_54}.

\bibitem[Hao et~al.(2015)Hao, Wang, Zhao, and Wang]{hao2015observer}
R.~Hao, J.~Wang, J.~Zhao, and S.~Wang.
\newblock Observer-based robust control of 6-dof parallel electrical manipulator with fast friction estimation.
\newblock \emph{IEEE Transactions on Automation Science and Engineering}, 13\penalty0 (3):\penalty0 1399--1408, 2015.

\bibitem[Hirose and Tajima(2017)]{Hirose2017}
N.~Hirose and R.~Tajima.
\newblock Modeling of rolling friction by recurrent neural network using lstm.
\newblock In \emph{2017 IEEE International Conference on Robotics and Automation (ICRA)}, pages 6471--6478, 2017.
\newblock \doi{10.1109/ICRA.2017.7989764}.

\bibitem[Huang and Tan(2012)]{Huang2012}
S.~Huang and K.~K. Tan.
\newblock Intelligent friction modeling and compensation using neural network approximations.
\newblock \emph{IEEE Transactions on Industrial Electronics}, 59\penalty0 (8):\penalty0 3342--3349, 2012.
\newblock \doi{10.1109/TIE.2011.2160509}.

\bibitem[Iskandar and Wolf(2019)]{iskandar2019}
M.~Iskandar and S.~Wolf.
\newblock Dynamic friction model with thermal and load dependency: modeling, compensation, and external force estimation.
\newblock In \emph{2019 International Conference on Robotics and Automation (ICRA)}, pages 7367--7373. IEEE, 2019.

\bibitem[Iskandar et~al.()Iskandar, Ott, Eiberger, Keppler, Albu-Sch{\"a}ffer, and Dietrich]{iskandar2020}
M.~Iskandar, C.~Ott, O.~Eiberger, M.~Keppler, A.~Albu-Sch{\"a}ffer, and A.~Dietrich.
\newblock Joint-level control of the dlr lightweight robot sara.
\newblock In \emph{2020 IEEE/RSJ International Conference on Intelligent Robots and Systems (IROS)}, pages 8903--8910. IEEE.

\bibitem[Iskandar et~al.(2021)Iskandar, Eiberger, Albu-Sch{\"a}ffer, De~Luca, and Dietrich]{iskandar2021}
M.~Iskandar, O.~Eiberger, A.~Albu-Sch{\"a}ffer, A.~De~Luca, and A.~Dietrich.
\newblock Collision detection, identification, and localization on the dlr sara robot with sensing redundancy.
\newblock In \emph{2021 IEEE International Conference on Robotics and Automation (ICRA)}, pages 3111--3117. IEEE, 2021.

\bibitem[Iskandar et~al.(2022)Iskandar, van Ommeren, Wu, Albu-Sch{\"a}ffer, and Dietrich]{iskandar2022}
M.~Iskandar, C.~van Ommeren, X.~Wu, A.~Albu-Sch{\"a}ffer, and A.~Dietrich.
\newblock Model predictive control applied to different time-scale dynamics of flexible joint robots.
\newblock \emph{IEEE Robotics and Automation Letters}, 8\penalty0 (2):\penalty0 672--679, 2022.

\bibitem[Iskandar et~al.(2023)Iskandar, Ott, Albu-Sch{\"a}ffer, Siciliano, and Dietrich]{iskandar2023hybrid}
M.~Iskandar, C.~Ott, A.~Albu-Sch{\"a}ffer, B.~Siciliano, and A.~Dietrich.
\newblock Hybrid force-impedance control for fast end-effector motions.
\newblock \emph{IEEE Robotics and Automation Letters}, 2023.

\bibitem[Johanastrom and Canudas-De-Wit(2008)]{johanastrom2008revisiting}
K.~Johanastrom and C.~Canudas-De-Wit.
\newblock Revisiting the lugre friction model.
\newblock \emph{IEEE Control systems magazine}, 28\penalty0 (6):\penalty0 101--114, 2008.

\bibitem[Jumper et~al.(2021)Jumper, Evans, Pritzel, Green, Figurnov, Ronneberger, Tunyasuvunakool, Bates, Z{\'i}dek, Potapenko, Bridgland, Meyer, Kohl, Ballard, Cowie, Romera-Paredes, Nikolov, Jain, Adler, Back, Petersen, Reiman, Clancy, Zielinski, Steinegger, Pacholska, Berghammer, Bodenstein, Silver, Vinyals, Senior, Kavukcuoglu, Kohli, and Hassabis]{Jumper2021}
J.~M. Jumper, R.~Evans, A.~Pritzel, T.~Green, M.~Figurnov, O.~Ronneberger, K.~Tunyasuvunakool, R.~Bates, A.~Z{\'i}dek, A.~Potapenko, A.~Bridgland, C.~Meyer, S.~A.~A. Kohl, A.~Ballard, A.~Cowie, B.~Romera-Paredes, S.~Nikolov, R.~Jain, J.~Adler, T.~Back, S.~Petersen, D.~A. Reiman, E.~Clancy, M.~Zielinski, M.~Steinegger, M.~Pacholska, T.~Berghammer, S.~Bodenstein, D.~Silver, O.~Vinyals, A.~W. Senior, K.~Kavukcuoglu, P.~Kohli, and D.~Hassabis.
\newblock Highly accurate protein structure prediction with alphafold.
\newblock \emph{Nature}, 596:\penalty0 583 -- 589, 2021.

\bibitem[Ke et~al.(2023)Ke, Yu, Li, Wang, Zhong, Wang, Kong, Guo, Huang, Idir, Liu, and Wang]{Ke2023}
X.~Ke, Y.~Yu, K.~Li, T.~Wang, B.~Zhong, Z.~Wang, L.~Kong, J.~Guo, L.~Huang, M.~Idir, C.~Liu, and C.~Wang.
\newblock Review on robot-assisted polishing: Status and future trends.
\newblock \emph{Robotics and Computer-Integrated Manufacturing}, 80:\penalty0 102482, 2023.
\newblock ISSN 0736-5845.
\newblock \doi{https://doi.org/10.1016/j.rcim.2022.102482}.

\bibitem[{Kemal Cılız} and Tomizuka(2007)]{Ciliz2007}
M.~{Kemal Cılız} and M.~Tomizuka.
\newblock Friction modelling and compensation for motion control using hybrid neural network models.
\newblock \emph{Engineering Applications of Artificial Intelligence}, 20\penalty0 (7):\penalty0 898--911, 2007.
\newblock ISSN 0952-1976.
\newblock \doi{https://doi.org/10.1016/j.engappai.2006.12.007}.

\bibitem[Kermani et~al.(2007)Kermani, Patel, and Moallem]{kermani2007friction}
M.~R. Kermani, R.~V. Patel, and M.~Moallem.
\newblock Friction identification and compensation in robotic manipulators.
\newblock \emph{IEEE Transactions on Instrumentation and Measurement}, 56\penalty0 (6):\penalty0 2346--2353, 2007.

\bibitem[Khan et~al.(2017)Khan, Chacko, and Nazir]{khan2017review}
Z.~A. Khan, V.~Chacko, and H.~Nazir.
\newblock A review of friction models in interacting joints for durability design.
\newblock \emph{Friction}, 5:\penalty0 1--22, 2017.

\bibitem[Kim et~al.(2021)Kim, Peternel, Lorenzini, Babi{\v{c}}, and Ajoudani]{kim2021human}
W.~Kim, L.~Peternel, M.~Lorenzini, J.~Babi{\v{c}}, and A.~Ajoudani.
\newblock A human-robot collaboration framework for improving ergonomics during dexterous operation of power tools.
\newblock \emph{Robotics and Computer-Integrated Manufacturing}, 68:\penalty0 102084, 2021.

\bibitem[Kingma and Ba(2015)]{Kingma2015Ba}
D.~P. Kingma and J.~Ba.
\newblock Adam: {A} method for stochastic optimization.
\newblock In Y.~Bengio and Y.~LeCun, editors, \emph{3rd International Conference on Learning Representations, {ICLR} 2015, San Diego, CA, USA, May 7-9, 2015, Conference Track Proceedings}, 2015.

\bibitem[Krizhevsky et~al.(2012)Krizhevsky, Sutskever, and Hinton]{Krizhevsky2012}
A.~Krizhevsky, I.~Sutskever, and G.~E. Hinton.
\newblock Imagenet classification with deep convolutional neural networks.
\newblock In F.~Pereira, C.~Burges, L.~Bottou, and K.~Weinberger, editors, \emph{Advances in Neural Information Processing Systems}, volume~25. Curran Associates, Inc., 2012.

\bibitem[Linderoth et~al.(2013)Linderoth, Stolt, Robertsson, and Johansson]{Linderoth2013}
M.~Linderoth, A.~Stolt, A.~Robertsson, and R.~Johansson.
\newblock Robotic force estimation using motor torques and modeling of low velocity friction disturbances.
\newblock In \emph{2013 IEEE/RSJ International Conference on Intelligent Robots and Systems}, pages 3550--3556, 2013.
\newblock \doi{10.1109/IROS.2013.6696862}.

\bibitem[Liu et~al.(2024)Liu, Rocco, Zanchettin, Zhao, Jiang, and Mei]{LIU2024102666}
B.~Liu, P.~Rocco, A.~M. Zanchettin, F.~Zhao, G.~Jiang, and X.~Mei.
\newblock A real-time hierarchical control method for safe human–robot coexistence.
\newblock \emph{Robotics and Computer-Integrated Manufacturing}, 86:\penalty0 102666, 2024.
\newblock ISSN 0736-5845.
\newblock \doi{https://doi.org/10.1016/j.rcim.2023.102666}.

\bibitem[Liu et~al.(2021)Liu, Wang, and Wang]{Liu2021}
S.~Liu, L.~Wang, and X.~V. Wang.
\newblock Sensorless force estimation for industrial robots using disturbance observer and neural learning of friction approximation.
\newblock \emph{Robotics and Computer-Integrated Manufacturing}, 71:\penalty0 102168, 2021.
\newblock ISSN 0736-5845.
\newblock \doi{https://doi.org/10.1016/j.rcim.2021.102168}.

\bibitem[P.~Aivaliotis and Makris(2023)]{aivaliotis2023}
D.~K.-G. P.~Aivaliotis and S.~Makris.
\newblock Physics-based modelling of robot’s gearbox including non-linear phenomena.
\newblock \emph{International Journal of Computer Integrated Manufacturing}, 36\penalty0 (12):\penalty0 1864--1875, 2023.
\newblock \doi{10.1080/0951192X.2022.2162594}.
\newblock URL \url{https://doi.org/10.1080/0951192X.2022.2162594}.

\bibitem[Reinecke et~al.(2014)Reinecke, Dietrich, Schmidt, and Chalon]{reinecke2014experimental}
J.~Reinecke, A.~Dietrich, F.~Schmidt, and M.~Chalon.
\newblock Experimental comparison of slip detection strategies by tactile sensing with the biotac{\textregistered} on the dlr hand arm system.
\newblock In \emph{2014 IEEE international Conference on Robotics and Automation (ICRA)}, pages 2742--2748. IEEE, 2014.

\bibitem[Rizos and Fassois(2005)]{Rizos2005}
D.~D. Rizos and S.~D. Fassois.
\newblock Friction identification based upon the lugre and maxwell slip models.
\newblock \emph{IFAC Proceedings Volumes}, 38\penalty0 (1):\penalty0 548--553, 2005.
\newblock ISSN 1474-6670.
\newblock \doi{https://doi.org/10.3182/20050703-6-CZ-1902.00092}.
\newblock 16th IFAC World Congress.

\bibitem[Selmic and Lewis(2002)]{Selmic2002}
R.~Selmic and F.~Lewis.
\newblock Neural-network approximation of piecewise continuous functions: application to friction compensation.
\newblock \emph{IEEE Transactions on Neural Networks}, 13\penalty0 (3):\penalty0 745--751, 2002.
\newblock \doi{10.1109/TNN.2002.1000141}.

\bibitem[Silver et~al.(2016)Silver, Huang, Maddison, Guez, Sifre, Driessche, Schrittwieser, Antonoglou, Panneershelvam, Lanctot, Dieleman, Grewe, Nham, Kalchbrenner, Sutskever, Lillicrap, Leach, Kavukcuoglu, Graepel, and Hassabis]{Silver2016}
D.~Silver, A.~Huang, C.~Maddison, A.~Guez, L.~Sifre, G.~Driessche, J.~Schrittwieser, I.~Antonoglou, V.~Panneershelvam, M.~Lanctot, S.~Dieleman, D.~Grewe, J.~Nham, N.~Kalchbrenner, I.~Sutskever, T.~Lillicrap, M.~Leach, K.~Kavukcuoglu, T.~Graepel, and D.~Hassabis.
\newblock Mastering the game of go with deep neural networks and tree search.
\newblock \emph{Nature}, 529:\penalty0 484--489, 01 2016.
\newblock \doi{10.1038/nature16961}.

\bibitem[Tu et~al.(2019)Tu, Zhou, Zhao, and Cheng]{tu2019modeling}
X.~Tu, Y.~Zhou, P.~Zhao, and X.~Cheng.
\newblock Modeling the static friction in a robot joint by genetically optimized bp neural network.
\newblock \emph{Journal of Intelligent \& Robotic Systems}, 94:\penalty0 29--41, 2019.

\bibitem[Vapnik(2000)]{Vapnik2000}
V.~Vapnik.
\newblock \emph{The Nature of Statistical Learning Theory}.
\newblock Springer: New York, 2000.

\bibitem[Vogel et~al.(2020)Vogel, Leidner, Hagengruber, Panzirsch, Bauml, Denninger, Hillenbrand, Suchenwirth, Schmaus, Sewtz, et~al.]{vogel2020ecosystem}
J.~Vogel, D.~Leidner, A.~Hagengruber, M.~Panzirsch, B.~Bauml, M.~Denninger, U.~Hillenbrand, L.~Suchenwirth, P.~Schmaus, M.~Sewtz, et~al.
\newblock An ecosystem for heterogeneous robotic assistants in caregiving: Core functionalities and use cases.
\newblock \emph{IEEE Robotics \& Automation Magazine}, 28\penalty0 (3):\penalty0 12--28, 2020.

\bibitem[Wolf and Iskandar(2018)]{wolf2018}
S.~Wolf and M.~Iskandar.
\newblock Extending a dynamic friction model with nonlinear viscous and thermal dependency for a motor and harmonic drive gear.
\newblock In \emph{2018 IEEE International Conference on Robotics and Automation (ICRA)}, pages 783--790. IEEE, 2018.

\bibitem[Xie et~al.(2023)Xie, Chong, Liu, Zhao, and Wang]{Xie2023}
F.~Xie, Z.~Chong, X.-J. Liu, H.~Zhao, and J.~Wang.
\newblock Precise and smooth contact force control for a hybrid mobile robot used in polishing.
\newblock \emph{Robotics and Computer-Integrated Manufacturing}, 83:\penalty0 102573, 2023.
\newblock ISSN 0736-5845.
\newblock \doi{https://doi.org/10.1016/j.rcim.2023.102573}.

\end{thebibliography}
\end{document}